# Structural Learning and Integrative Decomposition of Multi-View Data

Irina Gaynanova[1] and Gen Li[2]

[1]Department of Statistics, Texas A&M University
[2]Department of Biostatistics, Mailman School of Public Health, Columbia University

**Abstract**

The increased availability of the multi-view data (data on the same samples from multiple sources) has led to strong interest in models based on low-rank matrix factorizations. These models represent each data view via shared and individual components, and have been successfully applied for exploratory dimension reduction, association analysis between the views, and further learning tasks such as consensus clustering. Despite these advances, there remain significant challenges in modeling partially-shared components, and identifying the number of components of each type (shared/partially-shared/individual). In this work, we formulate a novel linked component model that directly incorporates partially-shared structures. We call this model SLIDE for Structural Learning and Integrative DEcomposition of multi-view data. We prove the existence of SLIDE decomposition and explicitly characterize the identifiability conditions. The proposed model fitting and selection techniques allow for joint identification of the number of components of each type, in contrast to existing sequential approaches. In our empirical studies, SLIDE demonstrates excellent performance in both signal estimation and component selection. We further illustrate the methodology on the breast cancer data from The Cancer Genome Atlas repository.

## 1 Introduction

Recent technological advances and elevated appreciation of the systems biology approach in biomedical fields has lead to increased availability of data from multiple sources. In cases where these data are collected on the same set of subjects, they are often called multi-view or multi-modal data. An illustrative example is The Cancer Genome Atlas (TCGA) repository, which contains data for the same set of subjects on genotype, gene expression, methylation expression, etc. One of the main challenges associated with the analysis of multi-view data is heterogeneity of measurements between the sources, which often leads to a separate analysis. A joint analysis, on the other hand, is essential for understanding the relationships between the sources, and moreover, has the potential to uncover biologically meaningful patterns beyond what is possible with separate approaches.

To address the challenge of heterogeneity in the joint analysis of multi-view data, a lot of methodological research has focused on linked component models. On a high level, these models take advantage of low-rank matrix factorizations to represent each source-specific dataset via shared and (possibly) individual latent components. The shared components can be used to study the associations between the views, or as inputs for further learning tasks such as consensus clustering (Lock and Dunson, 2013; Mo et al., 2013). The individual components can be used to study source-specific patterns of variability after adjusting for other sources. The latent components can be treated as either fixed or random leading to deterministic or probabilistic matrix factorizations respectively. Some examples are SUM-PCA (Smilde and Westerhuis, 2003), OnPLS (Löfstedt and Trygg, 2011), JIVE (Lock et al., 2013), iNMF (Yang and Michailidis, 2015), GFA (Klami et al., 2015) and COBE (Zhou et al., 2016). We refer the reader to Van Deun and Smilde (2009) and Zhou et al. (2016) for the review of deterministic matrix factorization models, and to Ray et al. (2014) for the review of bayesian models.

Despite these considerable methodological developments, there remain significant challenges in learning the structural factorization of multi-view data. A major challenge is associated with modeling partially-shared structures. For example, two views can share a latent component that is not present in the third view. Most existing models, however, assume that latent components are either globally shared or individual (Archambeau and Bach, 2009; Lock et al., 2013; Yang and Michailidis, 2015; Zhou et al., 2016). Our first contribution is the formulation of a novel deterministic linked component model that explicitly incorporates partially-shared structures.

While we are not the first to develop the framework for learning partially-shared structures, our approach and theoretical justifications are distinctive from the existing methods. Klami et al. (2015) consider probabilistic rather than deterministic matrix factorizations, and propose to use the hierarchical model with the structural sparsity prior. Their approach, however, requires some manual tuning of the parameters in prior specification. Jia et al. (2010) and Van Deun et al. (2011) also consider deterministic matrix factorizations, however neither existence, nor identifiability of the underlying models are discussed. In contrast, we provide a rigorous proof of existence of the proposed matrix decomposition for any given signal, and explicitly characterize the conditions that guarantee model uniqueness.

Another considerable challenge is associated with determining the number of components of each type (shared, individual, partially-shared) for each view. On the one hand, we are not aware of any principled approach for direct identification of ranks for partially-shared structures. On the other hand, even when only shared and individual structures are present, a common strategy is to first separately determine the total number of components for each view (total rank of each dataset), and only then divide them into shared and individual (Jia et al., 2010; Van Deun et al., 2011; Löfstedt and Trygg, 2011; Zhou et al., 2016; Feng et al., 2017). Given the presence of shared components, it is likely that the combination of best view-specific ranks is not the same as the best joint combination of ranks. JIVE (Lock et al., 2013) partially overcomes this problem by first selecting the number of shared components, and then selecting the number of individual components conditionally on the shared. The sequential approach, however, may still fail to identify the best joint combination. Given these drawbacks, our second contribution is the development of model selection approach that directly chooses the joint combination of the number of components of each type (shared, individual, partially-shared).

In summary, in this work we focus on deterministic matrix decomposition for multi-view data, and our main contributions are: (i) formulation of a novel deterministic linked component model that directly incorporates shared, partially-shared and individual structures, (ii) proof of existence of the proposed matrix decomposition together with the explicit characterization of the identifiabil-ity conditions, and (iii) derivation of model fitting and selection techniques for joint identification of the number of latent components for each type of structure. We call the proposed approach SLIDE for Structural Learning and Integrative DEcomposition of multi-view data.

The SLIDE model takes into account partially shared and individual components by exploiting structured sparsity in the matrix decomposition. We determine the total number of possible models, and consequently demonstrate the computational prohibitiveness of exhaustive search procedure for model selection. As a remedy, we propose to use penalized matrix factorization framework to reduce the model space to a small set of block-sparsity patterns for consideration, and adapt the bi-cross-validation procedure (Owen and Perry, 2009) to perform model selection out of this reduced set. Consequently, this leads to selection of the joint combination of ranks for shared, partially-shared and individual components. Finally, we develop a computationally efficient algorithm to fit the SLIDE model for the selected combination.

## 1.1 Notation

For a vector $v \in \mathbb{R}^p$, we use $\|v\|_2$ to denote the Eucledean norm. For a matrix $M \in \mathbb{R}^{n \times p}$, we use $m_{ij}$ to denote the elements of $M$, $\|M\|_F = \sqrt{\sum_{i=1}^n \sum_{j=1}^p m_{ij}^2}$ to denote the Frobenius norm, and $\|M\|_2 = \max_k \sigma_k(M)$ to denote the spectral norm with $\sigma_k(M)$ being the $k$th singular value of $M$. We use $\sigma_{\max}(M)$ to denote the largest singular value of $M$. We use $\text{col}(M)$ to denote the column space of matrix $M$, and $\text{row}(M)$ to denote the row space. We write $[M_1\ M_2] \in \mathbb{R}^{n \times (p_1 + p_2)}$ to denote the matrix formed by concatenating matrices $M_1 \in \mathbb{R}^{n \times p_1}$ and $M_2 \in \mathbb{R}^{n \times p_2}$ columnwise. If $M_2$ has only zero elements, we write $[M_1\ 0]$, and throughout the manuscript use 0 to denote the zero matrices of compatible size. We use $I_r = I \in \mathbb{R}^{r \times r}$ to denote the identity matrix of size $r$

## 1.2 Paper Organization

The rest of the manuscript is organized as follows. In Section 2 we formulate the SLIDE model together with the proof of existence and identifiability conditions, and discuss connections with other linked components models from the literature. In Section 3 we elaborate on SLIDE model selection and fitting procedure. In Section 4 we compare the performance of SLIDE with other competitors using simulated data. In Section 5 we use SLIDE for integrative analysis of multi-view data on breast cancer patients available from The Cancer Genome Atlas project. We conclude with discussion in Section 6.

## 2 Proposed Model

We consider $d$ datasets $X_i \in \mathbb{R}^{n \times p_i}$, $i = 1, \ldots, d$ from $d$ measurement sources on $n$ matched samples. We let $p = \sum_{i=1}^d p_i$ denote the total number of measurements across all sources. Following standard pre-processing (Lock et al., 2013; Smilde and Westerhuis, 2003; Van Deun et al., 2011), we column-center each dataset $X_i$ and standardize it so that $\|X_i\|_F = 1$. Our goal is to find a low-rank representation of each $X_i$ that allows the investigation of shared and individual signals across different views. We first review some existing linked component models that are based on extensions of the principal component analysis.

SUM-PCA(Smilde and Westerhuis, 2003; Van Deun and Smilde, 2009) considers the additive

noise model of rank $r$

$$X_i = UV_i^\top + E_i, \quad i = 1, \ldots, d, \qquad (1)$$

where $U \in \mathbb{R}^{n \times r}$ is the matrix of shared scores, $U^\top U = I_r$, $V_i \in \mathbb{R}^{p_i \times r}$ is the matrix of dataset-specific loadings, and $E_i \in \mathbb{R}^{n \times p_i}$ is the noise perturbation matrix. By forming the concatenated data matrix $X = [X_1 \ldots X_d] \in \mathbb{R}^{n \times p}$, the concatenated loadings matrix $V = [V_1^\top \ldots V_d^\top]^\top \in \mathbb{R}^{p \times r}$ and the concatenated error matrix $E = [E_1 \ldots E_d]$, model (1) can be rewritten as

$$X = UV^\top + E, \qquad (2)$$

which reveals that the score vectors in $U$ are shared across all $d$ datasets, and moreover, $U$ and $V$ can be found by performing PCA on concatenated data matrix $X$. A major limitation of SUM-PCA is that it does not allow for partially-shared or individual scores.

Alternatively, performing rank $r_i$ PCA on each $i$th dataset individually corresponds to the model

$$X_i = U_i V_i^\top + E_i, \quad i = 1, \ldots, d, \qquad (3)$$

where now each $U_i \in \mathbb{R}^{n \times r_i}$, $U_i^\top U_i = I_{r_i}$, is the matrix of individual scores, and therefore there is no sharing of information across the datasets. JIVE (Lock et al., 2013) can be viewed as an intermediate model between (1) and (3) as it corresponds to

$$X_i = U_0 V_{0i}^\top + U_i V_i^\top + E_i, \quad i = 1, \ldots, d, \qquad (4)$$

where score vectors in $U_0$ are shared across all datasets, and score vectors in $U_i$ are dataset-specific so that $U_0^\top U_i = 0$. JIVE model, however, does not allow for partially-shared scores. Moreover, the individual matrices do not have to be mutually orthogonal, which requires care in their interpretation as individual as discussed in Section 2.1.

In this work, we propose a direct generalization of model (2) by considering structured sparsity of loadings matrix $V \in \mathbb{R}^{p \times r}$. Specifically, we divide $V$ row-wise into blocks $V_1, \ldots, V_d$ corresponding to $d$ datasets, and allow some of the columns in data-specific blocks $V_i$ to be exactly zero. The resulting block-sparse structure of $V$ parsimoniously takes into account shared, partially-shared and individual scores. For example, when $d = 3$, each column of $V$ corresponds to one of the 8 sparsity patterns:

$$V = \begin{pmatrix} V_1 \\ V_2 \\ V_3 \end{pmatrix} = \begin{pmatrix} V_{1,1} & V_{1,2} & V_{1,3} & 0 & V_{1,5} & 0 & 0 & 0 \\ V_{2,1} & V_{2,2} & 0 & V_{2,4} & 0 & V_{2,6} & 0 & 0 \\ V_{3,1} & 0 & V_{3,3} & V_{3,4} & 0 & 0 & V_{3,7} & 0 \end{pmatrix},$$

where each $V_{i,k} \in \mathbb{R}^{p_i \times r_k}$ with $p_i$ being the number of measurements in dataset $i$, $i = 1, 2, 3$, and $r_k$ being the number of columns with $k$th sparsity pattern, $k = 1, \ldots, 8$.

Substituting the above $V$ in model (1), and writing $U = [U_1 \ldots U_7]$ leads to each $X_i$ being expressed as

$$\begin{aligned} X_1 &= U_1 V_{1,1}^\top + U_2 V_{1,2} + U_3 V_{1,3} &&+ U_5 V_{1,5}^\top + E_1, \\ X_2 &= U_1 V_{2,1}^\top + U_2 V_{2,2} &&+ U_4 V_{2,4} + U_6 V_{2,6}^\top + E_2, \\ X_3 &= U_1 V_{3,1}^\top &&+ U_3 V_{3,3} + U_4 V_{3,4} + U_7 V_{3,7}^\top + E_3, \end{aligned}$$

where scores $U_1$ are shared across all datasets, scores $U_2$ are shared across datasets 1 and 2, scores $U_3$ are shared across datasets 1 and 3, scores $U_4$ are shared across datasets 2 and 3, and scores $U_5$, $U_6$, $U_7$ are unique to datasets 1, 2 and 3 respectively. Therefore, block-sparsity of matrix $V$ allows the identification of partially-shared and individual scores. Moreover, the number of non-zero columns in $V(r = \sum_{k=1}^{7} r_k)$ corresponds to the total number of components for all datasets, whereas each $r_k$, $k = 1, ..., 8$, corresponds to the number of components in accordance with the $k$th sparsity pattern. For convenience, we use a binary matrix $S \in \{0,1\}^{d \times r}$ to represent the block-sparse structure of matrix $V \in \mathbb{R}^{\sum_{i=1}^{d} p_i \times r}$ so that $s_{ij} = 0$ if $j$th column of $i$th data block $V_i$ is zero, and $s_{ij} = 1$ if it is non-zero. Next, we formally state the proposed model.

We represent the concatenated data matrix $X \in \mathbb{R}^{n \times p}$ of centered and standardized $X_i$ using a binary matrix $S \in \{0,1\}^{d \times r}$, a score matrix $U \in \mathbb{R}^{n \times r}$ with $U^\top U = I_r$, a concatenated loading matrix $V = V(S) \in \mathbb{R}^{p \times r}$ with block-sparsity pattern according to $S$, and a noise perturbation matrix $E$ such that

$$X = UV(S)^\top + E. \tag{5}$$

We call this model SLIDE for Structural Learning and Integrative DEcomposition of multi-view data.

## 2.1 Connections to Other Linked Component Models

If all elements of the binary matrix $S \in \{0,1\}^{d \times r}$ are equal to one, then SLIDE model (5) reduces to SUM-PCA model (2) with $r$ score vectors. If each column of the binary matrix $S \in \{0,1\}^{d \times r}$ has exactly one non-zero element, all the corresponding scores are individual, and SLIDE model (5) reduces to individual PCA (3) with additional requirement of orthogonality between individual scores. If each column of the binary matrix $S \in \{0,1\}^{d \times r}$ is either equal to the vector of ones $\mathbf{1} = \{1\}^d$ or has exactly one non-zero element (this is always true when $d = 2$), then SLIDE model (5) reduces to JIVE model (4) with additional requirement of orthogonality between individual scores. Therefore, SLIDE encompasses several existing linked component models as special cases.

In contrast to the existing approaches, individual score vectors are mutually orthogonal in SLIDE. The orthogonality allows to both identify the number of shared components from the canonical correlation analysis perspective, and interpret individual part of the decomposition as the view-specific signal that remains after accounting for other sources. This is not the case for non-orthogonal components as illustrated in the following toy example. Moreover, the orthogonality significantly simplifies the model fitting procedure in Section 3.

Consider two matched datasets expressed through their rank-one individual PCA decomposition (3) as

$$X_1 = u_1 v_1^\top + E_1; \quad X_2 = u_2 v_2^\top + E_2 \quad \text{with} \quad u_1^\top u_1 = u_2^\top u_2 = 1. \tag{6}$$

Further, let $u_1^\top u_2 = c \in (0, 1)$. Without the orthogonality requirement, $u_1$ and $u_2$ would be considered individual scores since $u_1 \neq u_2$. On the other hand, the canonical correlation analysis applied to $Z_1 = u_1 v_1^\top$ and $Z_2 = u_2 v_2^\top$ reveals that the canonical correlation is equal to $c$. In case $c$ is large, this indicates a strong association between $X_1$ and $X_2$, despite the absence of shared components.

In contrast, $X_1$ and $X_2$ can be decomposed using SLIDE model (5) as:

$$X_1 = u_1 v_1^\top + E_1; \quad X_2 = u_1 c v_2^\top + (I - u_1 u_1^\top) u_2 v_2^\top + E_2 = u_1 \widetilde{v}_{2,1}^\top + \widetilde{u}_2 \widetilde{v}_{2,2}^\top + E_2, \tag{7}$$

with corresponding structure matrix $S = \begin{pmatrix} 1 & 0 \\ 1 & 1 \end{pmatrix}$. The score $u_1$ is shared across both datasets, correctly indicating the presence of association. Moreover, since $\widetilde{u}_2$ is orthogonal to $u_1$, the canonical correlation is zero between $u_1 v_1^\top$ and $\widetilde{u}_2 \widetilde{v}_{2,2}^\top$, hence variation in $\widetilde{u}_2 \widetilde{v}_{2,2}^\top$ can not be explained by $X_1$. The decomposition (7), however, is not the only decomposition satisfying the SLIDE model, and an alternative with $S = \begin{pmatrix} 1 & 1 \\ 1 & 0 \end{pmatrix}$ can be considered. Despite this non-uniqueness, both SLIDE representations identify the presence of rank one shared structure, in contrast to (6). We discuss the existence and uniqueness of SLIDE model in the next section.

## 2.2 Model Existence and Identifiability

First, we discuss the equivalence between different binary matrices $S \in \{0,1\}^{d \times r}$ in terms of specifying the block-sparsity pattern in SLIDE model (5). If $S$ has zero columns, then the corresponding columns of $V(S)$ are zero, and deleting these columns leads to an equivalent decomposition. Similarly, an equivalent decomposition can be obtained by permuting the columns of $S$ together with the corresponding columns of $U$ and $V(S)$. We formally characterize the equivalence relationships between different structure matrices $S$ below.

**Definition 1.** *Two matrices $S_1 \in \{0,1\}^{d \times r_1}$ and $S_2 \in \{0,1\}^{d \times r_2}$ with $r_1 \geq r_2$ give rise to an equivalent SLIDE decomposition (5) if there exists an $r_1 \times r_1$ permutation matrix $P_\pi$ corresponding to permutation $\pi : \{1, \ldots, r_1\} \to \{1, \ldots, r_1\}$ such that*

$$S_1 P_\pi = \widetilde{S}_2,$$

*where $\widetilde{S}_2 \in \{0,1\}^{d \times r_1}$ is formed from $S_2$ by appending $r_1 - r_2$ zero columns.*

Therefore, two structure matrices $S_1$ and $S_2$ are equivalent if they are equal up to the permutation of their columns, and addition/deletion of zero columns.

Next, we discuss the existence of SLIDE decomposition (5), and its uniqueness according to Definition 1. For this purpose, we let $\mathcal{B}_d$ define the set of distinct non-zero binary vectors $b_i \in \{0,1\}^d$. Since there are $2^d - 1$ such vectors, we write $\mathcal{B}_d = \{b_1, \ldots, b_{2^d-1}\}$, for example $\mathcal{B}_2 = \{b_1 = (1,1), b_2 = (1,0), b_3 = (0,1)\}$. Each non-zero column of binary structure matrix $S \in \mathbb{R}^{d \times r}$ must be equal to one of the $2^d - 1$ distinct binary vectors $b_i \in \mathcal{B}_d$. In light of Definition 1, we assume without loss of generality that $S$ has only non-zero columns, and that the columns are sorted with respect to the values $\{b_1, \ldots, b_{2^d-1}\}$.

**Theorem 1.** *Consider matched datasets $X_i \in \mathbb{R}^{n \times p_i}$ with corresponding concatenated data matrix $X = [X_1 \ldots X_d] \in \mathbb{R}^{n \times p}$ that follows the additive noise model $X = Z + E$. Then*

1. *There exists a binary matrix $S \in \{0,1\}^{d \times r}$, a score matrix $U \in \mathbb{R}^{d \times r}$, $U^\top U = I$, and a loadings matrix $V = V(S) \in \mathbb{R}^{p \times r}$ with block-sparsity pattern according to $S$ such that $Z = UV(S)^\top$. Moreover, for each $b_k \in \mathcal{B}_d$, the corresponding non-zero columns of dataset-specific loadings $V_i$ are linearly independent.*

2. *Consider the SLIDE decomposition from part 1. If all non-zero columns of dataset-specific loadings matrix $V_i$ are linearly independent for each $i = 1, \ldots, d$, then $S$ is unique. In addition, if for each $b_k \in \mathcal{B}_d$, the corresponding columns of $V$ are orthogonal with distinct norms, then $U$ and $V$ are also unique.*

**Remark 1.** 1. *The linear independence property in part 1 ensures that for each $b_k \in \mathcal{B}_d$, the rank of corresponding submatrix of $V_i$ is the same for each dataset $i$ with $b_{ki} = 1$, and is equal* to the number of corresponding columns in $U$. This, in particular, distinguishes SLIDE from singular value decomposition on $Z$, and moreover, ensures the correct interpretation of the shared structure for each $b_k \in \mathcal{B}_d$.

2. The requirement of distinct column norms in part 2 is similar to the uniqueness requirements in the singular value decomposition. The columns of $V$ with the same norms and the same pattern $b_k \in \mathcal{B}_d$ are unique only up to an orthogonal rotation.

Theorem 1 asserts the existence of SLIDE decomposition and its uniqueness under the condition of linear independence of nonzero columns in dataset-specific loadings $V_i$. While this condition may not be satisfied for a given signal matrix $Z$, we found that it always holds for fitted SLIDE model due to the corruption of signal by noise. Further discussion regarding model identifiability, as well as comparison with identifiability conditions of JIVE model (Lock et al., 2013), can be found in the Appendix A.

## 3 SLIDE Structure Selection and Model Fitting

Our goal is to simultaneously learn the binary structure matrix $S$ (structural learning) as well as the score matrix $U$ and the loadings matrix $V = V(S)$ (integrative decomposition) in the SLIDE model (5). A direct approach for structural learning is to consider all distinct $S$, and choose the "best" according to some pre-specified procedure (for example percentage of variance explained). While the choice of selection procedure is crucial, and we discuss our proposed approach in Section 3.2, we want to emphasize that no procedure can do exhaustive search over all $S$ in polynomial time.

**Proposition 1.** *For a given $r$, consider the set $\mathcal{S}_r$ of binary matrices $S \in \{0,1\}^{d \times r}$. Then the number of elements in $\mathcal{S}_r$ that are distinct according to Definition 1 is equal to*

$$\binom{r + 2^d - 1}{2^d - 1}.$$

**Remark 2.** *The number of columns $r$ in Proposition 1 represents the maximal possible rank of $UV(S)^\top$ in (5), which can be as large as the rank of $X$. In the high-dimensional settings, it is typical to have $rank(X) \approx \min(n,p) = n$, which leads to $\binom{n+2^d-1}{2^d-1}$ distinct $S$. In what follows, we always set $r = \min(n,p)$.*

Given Proposition 1, we consider the following challenges in this work:

1. How to reduce the number of structures for consideration from $\binom{r+2^d-1}{2^d-1}$ to a "small" subset? (structural learning)

2. Given a sequence of $m$ distinct binary structures $S_1, \ldots, S_m$, how to choose the "best"? (structural learning)

3. Given a binary structure $S \in \{0,1\}^{d \times r}$, how to fit the SLIDE model (5) based on the concatenated data matrix $X$? (integrative decomposition)

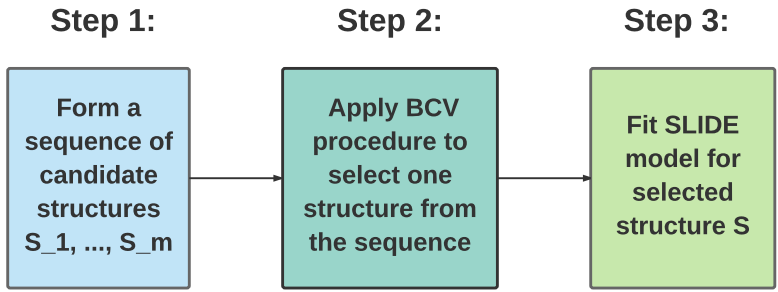

Figure 1: The summary of the proposed 3-step approach for SLIDE model fitting.

Consequently, we use penalized matrix factorization framework to generate a small set of candidate structure matrices $S_1, \ldots, S_m$ (Section 3.1). We adapt the bi-cross-validation (BCV) approach (Owen and Perry, 2009) to select the "best" structure from the fixed set (Section 3.2). We devise an iterative algorithm to fit SLIDE model (5) for a given structure $S$ (Section 3.3). The overall workflow of SLIDE framework is presented in Figure 1. The modularity of the proposed approach allows great flexibility. For example, one can use a different subset selection approach for Step 1 or explicitly include certain structures, and use Steps 2 and 3 without modification. Similarly, one can use a different structure selection algorithm in Step 2, but still use Step 1 to form the reduced structure sequence. Our choice of methods for Steps 1 and 2 is motivated by their computational efficiency, and excellent empirical performance (Section 4).

## 3.1 Reducing the Number of Structures for Consideration

Our first objective is to form a small candidate set our of the set of all possible structures. Ideally, we would like this reduced candidate set to include a wide range of structures $S$ varying from a small number of non-zero columns to a large number of non-zero columns. For this purpose, we use penalized matrix factorization framework, namely we combine the Frobenius norm loss with convex block-sparsity-inducing penalty on the columns of $V$, and consider

$$(\widetilde{U}, \widetilde{V}) = \operatorname*{argmin}_{U \in \mathbb{R}^{n \times r}, V \in \mathbb{R}^{p \times r}} \left\{ \sum_{i=1}^{d} \left[ \frac{1}{2} \|X_i - UV_i^\top\|_F^2 + \lambda \sum_{j=1}^{r} \|V_{ij}\|_2 \right] \right\} \quad \text{s.t} \quad U^\top U = I, \qquad (8)$$

where $V_{ij} \in \mathbb{R}^{p_i}$ is the $j$th column of the $i$th block $V_i \in \mathbb{R}^{p_i \times r}$ of loadings matrix $V$. We choose $r = \min(n, p)$. Large values of tuning parameter $\lambda$ lead to a large number of zero blocks in loadings matrix $V$, hence large number of zeroes in corresponding structure matrix $S$. In contrast, small values of $\lambda$ lead to small number of zeroes. Therefore, by considering a grid of tuning parameters from small to large in (8), we can generate a sequence of structure matrices $S$ with different complexity based on corresponding solutions $\widetilde{V}$. Moreover, the total number of resulting distinct structures is always upper bounded by the number of tuning parameters.

Given a sequence of tuning parameters $\lambda_1 < \cdots < \lambda_k$, we use the corresponding support of $\widetilde{V}$ to generate the sequence of structures $S_1, \ldots, S_k \in \{0,1\}^{d \times r}$ (note that any structure $S' \in \{0,1\}^{d \times r'}$

**Algorithm 1** Iterative algorithm for solving (8)
---
Given: $X = [X_1 \ldots X_d] \in \mathbb{R}^{n \times p}$, $U^{(0)}$, $\lambda \geq 0$, $\varepsilon > 0$
  $V^{(0)} \leftarrow X^\top U^{(0)}$
  $k \leftarrow 1$
  **repeat**
    Update of $V$:
      **for** $i = 1$ **to** $d$ **do**
        **for** $j = 1$ **to** $r$ **do**
$$V_{ij}^{(k)} \leftarrow \max\left(0, \left[1 - \frac{\lambda}{\|X_i^\top U_j^{(k-1)}\|_2}\right]\right) X_i^\top U_j^{(k-1)}$$
        **end for**
      **end for**

    Update of $U$:
      Singular value decomposition $XV^{(k)} = RLQ^\top$
      $U^{(k)} \leftarrow RQ^\top$
  $k \leftarrow k + 1$
  **until** $f(U^{(k-1)}, V^{(k-1)}) - f(U^{(k)}, V^{(k)}) < \varepsilon$
---

with $r' \leq r$ can always be represented as $S \in \{0,1\}^{d \times r}$ by appending $r - r'$ zero columns). Since neighboring tuning parameters $\lambda_i, \lambda_{i+1}$ may lead to equivalent structures, we further trim the list $S_1, \ldots, S_k$ to remove any repetitions. As a result, we get a sequence of distinct binary structures $S_1,\ldots,S_m$ with $m \leq k$, which serves as an input for Step 2 of SLIDE workflow (Figure 1).

To generate the sequence of $\lambda$ values, we use a logarithmic grid from 0.01 to $\lambda_{\max} = \max_i \sigma_{\max}(X_i)$ of length 50, where we choose $\lambda_{\max}$ according to the following proposition.

**Proposition 2.** *Define $\lambda_{\max}$ as the minimal value of $\lambda$ such that for all $\lambda \geq \lambda_{\max}$, $\widetilde{V} = 0$. Then*

$$\lambda_{\max} \leq \max_{1 \leq i \leq d} \sigma_{\max}(X_i),$$

*where $\sigma_{\max}(X_i)$ is the maximal singular value of $X_i$.*

Next, we discuss how to solve problem (8) for a fixed value of $\lambda \geq 0$. Let $f(U, V)$ denote the objective function in (8). We perform alternating minimization of $f(U, V)$ with respect to $U$ and $V$. Both updates can be performed in closed form leading to Algorithm 1, the full derivations can be found in the Appendix C. While problem (8) is nonconvex, the algorithm always converges to the partial optimum.

**Proposition 3.** *The algorithm 1 is guaranteed to converge. Moreover, let $f^* = f(U^*, V^*)$ be the function value at convergence, then*

$$f^* = \min_{U: U^\top U = I} f(U, V^*) = \min_V f(U^*, V),$$

*and $U^*, V^*$ are partial optimum.*

As problem (8) is nonconvex, convergence to the global optimum is not guaranteed, and the output of the algorithm depends on the starting point $U^{(0)}$. We found that initializing $U^{(0)}$ with the matrix of left singular vector of concatenated $X$ works well in the scenarios considered in Section 4. By Eckhart-Young theorem, this choice of $U^{(0)}$ corresponds to the global solution when $\lambda = 0$. For the dataset analysis of Section 5, we use random restarts to produce a more accurate list of structures.

Finally, we note that while a criterion similar to (8) has been previously used in Van Deun et al. (2011), here we only use (8) as a tool for identifying the set of candidate structures. We do not use (8) for SLIDE model-fitting for several reasons. First, the sparsity penalty introduces bias in the resulting estimates of $U$ and $V$, and it has been previously found in the context of sparse linear regression that the estimation error can be improved by refitting the model on the support (Efron et al., 2004; Meinshausen, 2007). Secondly, we found that this bias leads to selection of wrong structure when the choice of tuning parameter is based on minimizing the Frobenius norm error. Finally, the proposed BCV approach for model selection (Section 3.2) is tailored for structure identification rather than the Frobenius norm minimization. It can not be applied to the selection of tuning parameter in (8) since the same sequence of tuning parameters leads to different structures across folds.

## 3.2 Selecting the Structure via Bi-Cross-Validation

In this section, we adapt the BCV procedure of Owen and Perry (2009) to select the "best" structure $S$ out of the sequence of candidate structures $S_1, ..., S_m$ for SLIDE model (5).

We start by reviewing the BCV, which serves as an extension of k-fold cross-validation for the purpose of rank estimation. Given a matrix $X \in \mathbb{R}^{n \times p}$ and a sequence of candidate ranks $r_1, \ldots, r_m$, it aims to choose the "best" rank for $X$. For this purpose, BCV splits both the rows and the columns of $X$ into folds to form submatrices. For example, 2 folds over rows and 2 folds over columns lead to 4 submatrices as follows:

$$X = \begin{pmatrix} X^{11} & X^{12} \\ X^{21} & X^{22} \end{pmatrix}.$$

The BCV proceeds by holding out each submatrix at a time, and evaluating the Frobenius norm loss of the estimate which is obtained from the remaining submatrices with respective candidate ranks. More specifically, suppose $X^{11}$ is hold out. Then for each $l \in \{r_1, \ldots, r_m\}$, the method performs 2 steps:

1. Fit rank $l$ decomposition to $X^{22}$, for example use the $l$th truncated SVD of $X^{22}$ to form scores $U \in \mathbb{R}^{n \times l}$ and loadings $V \in \mathbb{R}^{p \times l}$ (other fits are possible, but we restrict presentation to SVD to fix the ideas)

2. Evaluate the error of the fit on $X^{11}$ through $X^{12}$ and $X^{21}$ as

$$\|X^{11} - X^{12}V(V^\top V)^{-1}(U^\top U)^{-1}U^\top X^{21}\|_F^2.$$

The intuition behind the error evaluation is that $X^{21}$ is used to predict the column structure of $X^{11}$ via $U$, and $X^{12}$ is used to predict the row structure of $X^{11}$ via $V$. In case more than two folds are considered, the submatrices that are not hold out are combined reducing to the case above. The best rank $l \in \{r_1, \ldots, r_m\}$ is chosen as the one that minimizes total error across all submatrices.

Owen and Perry (2009) demonstrate the superior empirical performance of BCV compared to other rank selection approaches, and in special cases provide theoretical guarantees on BCV performance.

The BCV procedure is designed to select a rank for individual matrix. In our case, however, we are interested in selecting a structure $S$ for SLIDE model that involves $d$ matrices. We propose the following adaptation of BCV procedure. Let $X_1, \ldots, X_d$ be the $d$ matched datasets. Let $k_r$ be the number of folds for rows across all $d$ datasets, and $k_c$ be the number of folds for columns within each dataset. For the ease of presentation, we illustrate the case $k_r = k_c = 2$, and $d = 2$ below:

$$X = [X_1 \ X_2] = \begin{pmatrix} X_1^{11} & X_1^{12} & X_2^{11} & X_2^{12} \\ X_1^{21} & X_1^{22} & X_2^{21} & X_2^{22} \end{pmatrix}.$$

Instead of holding one submatrix at a time from the whole dataset $X$, we hold out $d$ submatrices (one from each $X_i$ respectively). Moreover, we take into account centering and standardization of $X$ in SLIDE model (5) when obtaining the estimate and evaluating the prediction error. Suppose $X^{11} = [X_1^{11} \ldots X_d^{11}]$ is hold out. Then for each $S \in \{S_1, \ldots, S_m\}$, the adjusted BCV selection approach performs 3 steps:

1. Form $\widetilde{X}^{22} = [\widetilde{X}_1^{22} \ldots \widetilde{X}_d^{22}]$ by column centering and scaling $X^{22} = [X_1^{22} \ldots X_d^{22}]$ so that $\|\widetilde{X}_i^{22}\|_F = 1$. Fit the SLIDE model for $\widetilde{X}^{22}$ using Algorithm 2 in Section 3.3 with structure $S$ to find $U$ and $V$.

2. Perform back-scaling and back-centering by forming

$$\widehat{U} = \left(\frac{1}{\sqrt{n_r}}1, \ U\right) \quad \text{and} \quad \widehat{V} = \left(\frac{1}{\sqrt{n_r}}X^{22\top}1, \ V'\right),$$

where $1 \in \mathbb{R}^{n_r}$ is the vector of ones, $V'_i = V_i \|X_i^{22}\|_F$ for $i = 1, \ldots, d$, and $n_r$ is the number of rows in $X^{22}$.

3. Evaluate the prediction error on $X^{11} = [X_1^{11} \ldots X_d^{11}]$ via $[X_1^{12} \ldots X_d^{12}]$, $[X_1^{21} \ldots X_d^{21}]$, $\widehat{U}$ and $\widehat{V}$ as

$$\sum_{i=1}^d \frac{1}{\|(I - 11^\top)X_i^{11}\|_F^2} \|X_i^{11} - X^{12}\widehat{V}(\widehat{V}^\top \widehat{V})^{-1}\widehat{U}^\top X_i^{21}\|_F^2,$$

where $(I - 11^\top)X_i^{11}$ is the column-centered $X_i^{11}$. The above expression calculates Frobenius norm error on each dataset-specific block $X_i^{11}$ relative to the amount of total variance ($\|(I - 11^\top)X_i^{11}\|_F^2$), and then sums the scaled errors over all datasets. For each $X_i^{11}$, $X_i^{21}$ is used to predict the column structure (which is view-specific), and all $X^{12}$ is used to predict the row structure (which is shared between the datasets).

We select the structure that minimizes the total prediction error across all folds.

## 3.3 Fitting SLIDE with Selected $S$

In this section we describe the process of fitting the SLIDE model (5) for a given structure $S \in \{0, 1\}^{d \times r}$. Without loss of generality, we assume that $S$ has only non-zero columns. Similar to

**Algorithm 2** Iterative algorithm for fitting SLIDE model with a pre-specified structure $S$

---
Given: $X_i \in \mathbb{R}^{n \times p_i}$, $i = 1, ..., d$, $S \in \mathbb{R}^{r \times d}$, $U^{(0)} \in \mathbb{R}^{n \times r}$ such that $U^{(0)\top} U^{(0)} = I_r$, $\varepsilon > 0$

  $k \leftarrow 0$
  **repeat**
    Update of $V$:
      **for** $i = 1$ **to** $d$ **do**
        Determine the structure present in dataset $i$: $F \leftarrow \{l : s_{il} == 1\}$
        $V_{i,F} \leftarrow X_i^\top U_F$
        $V_{i,F^c} \leftarrow 0$
      **end for**

    Update of $U$:
      Singular value decomposition $XV^{(k)} = RLQ^\top$
      $U^{(k+1)} \leftarrow RQ^\top$
  $k \leftarrow k + 1$
  **until** $\left\| U^{(k)} V^{(k)\top} - U^{(k-1)} V^{(k-1)\top} \right\|_F^2 < \varepsilon$

  Orthogonalization for each $b_k \in \mathcal{B}_d$:
    Singular value decomposition on block corresponding to $b_k$: $U_{b_k} V_{b_k}^\top = RLQ^\top$
    $U_{b_k} \leftarrow R$, $V_{b_k} \leftarrow LQ^\top$

---

other linked component models, we fit SLIDE by minimizing the Frobenius norm error, that is by solving

$$\begin{aligned}
\operatorname*{minimize}_{U \in \mathbb{R}^{n \times r}, V \in \mathbb{R}^{p \times r}} & \quad \|X - UV^\top\|_F^2 \\
\text{subject to} & \quad U^\top U = I \\
& \quad V = V(S).
\end{aligned} \qquad (9)$$

When all elements of $S$ are equal to one, by Eckhart-Young Theorem the solution coincides with rank $r$ truncated SVD of concatenated $X$ (Golub and Van Loan, 2012, Theorem 2.4.8). When some elements of $S$ are equal to zero, we propose to use alternating minimization with respect to $U$ and $V$. From the numerical perspective, SVD is typically found by performing a series of iterative left and right-multiplications with an additional orthogonalization step. This motivates us to directly adjust the SVD iterations to fit the SLIDE model, details are presented in Algorithm 2. Each update of Algorithm 2 is exactly solving the optimization problem (9) with respect to only one variable ($U$ or $V$), and therefore is guaranteed to converge to partial optimum. The proof is similar to Proposition 3, and therefore is omitted. As with Algorithm 1, we initialize Algorithm 2 with $U^{(0)}$ formed from $r$ leading left singular vectors of $X$.

To ensure uniqueness of $U$ and $V$ according to Theorem 1, we perform an additional orthogonalization step within each sparsity pattern $b_k \in \mathcal{B}_d$ after the convergence of the algorithm. This orthogonalization step has no effect on the Frobenius norm error, or the block-sparsity pattern since it is performed independently for each block. Finally, we note that while (9) does not enforce the linear independence in the columns of $V_i$ as stated in Theorem 1, we found that this linear independence always holds in practice for the resulting output due to the corruption of signal by noise. Therefore, the resulting fitted SLIDE model satisfies the identifiability condition of Theorem 1.

## 4 Simulation Results

In this section, we investigate the performance of SLIDE using simulated data, and follow the steps in Figure 1. In Step 1, we use a grid of tuning parameters of length 50 on a logarithmic scale from 0.01 to $\lambda_{\max}$ from Proposition 2 to generate a sequence of binary structures $S_1, .., S_m$ for consideration. The best structure is selected according to BCV criterion described in Section 3.2 with 3 folds used for both rows and columns, and the final decomposition is found by applying Algorithm 2 with selected structure $S$.

For comparison, we also consider the performance of JIVE (Lock et al., 2013) as implemented in the R package **r.jive** (O'Connell and Lock, 2016) with default option of selecting the number of components using the permutation scheme. While there exist many other linked component models, we have chosen to compare with JIVE since: (1) it is based on deterministic matrix decomposition that allows for both shared and individual components, (2) it automatically determines the number of components of each type, (3) its implementation does not require user's input and (4) it has been proven to be a powerful tool for the analysis of multi-view data (Kuligowski et al., 2015; Hellton and Thoresen, 2016). In particular, (1) ensures that the simulated models are favorable to both methods, and (2)-(3) give us confidence that the results are not dependent on our ability to correctly "tune" the parameters. To our knowledge, this is not the case for many other methods which are either based on significantly different matrix factorizations, or require manual tuning by the user.

For all settings, we set $n = 100$, and generate $d$ matched datasets $X_i \in \mathbb{R}^{n \times p_i}$ as

$$X_i = Z_i + E_i,$$

where $Z_i \in \mathbb{R}^{n \times p_i}$ correspond to true signal, and $E_i$ are generated with independent entries such that $e_{i,kj} \sim N(0, \sigma_i^2)$, $i = 1, ..., d$. In each simulation, we encode shared, partially-shared and individual structures into $Z_i$, and set $\sigma_i^2$ such that

$$\|Z_i\|_F^2 / (\sigma_i^2 n p_i) = 1 \quad \text{for} \quad i = 1, \ldots, d. \tag{10}$$

The quantity in the denominator above is $\mathbb{E}(\|E_i\|_F^2)$ under assumptions of independence and normality.

To evaluate the performance of the methods in recovering the true signals $Z_i$, we compare the estimated ranks of shared, partially-shared and individual structures with the true ranks. We also evaluate the scaled squared Frobenius norm error defined as

$$L_F(Z, \widehat{Z}) = \sum_{i=1}^{d} \frac{1}{\|Z_i\|_F^2} \|Z_i - \widehat{Z}_i\|_F^2, \tag{11}$$

where $\widehat{Z}_i$ is the estimated signal for the $i$th dataset. Similar performance metric has been used in Yang and Michailidis (2015). We chose this metric as it is scale invariant, and furthermore, is independent from a particular decomposition used for $Z_i$, making it invariant with respect to the definitions of shared and individual signals. The latter property, in particular, allows us to objectively assess the performance of JIVE in the presence of partially-shared structures.

To understand the effect of structure selection procedure on the resulting comparison, we fit SLIDE with each given structure in the reduced set, and select the one with the smallest Frobenius norm difference from the true signal (**SLIDE_best**). We do not enforce the true structure to be explicitly included in the reduced set. We also use the true structure to directly fit JIVE model with correct number of components of each type (**JIVE_best**). Finally, we also consider a PCA-based decomposition given the true number of components where we first extract the shared structure using SVD on concatenated dataset $X = [X_1 \dots X_d]$, and then extract the individual structures on the residual individual datasets. We call this procedure **Onestep** to emphasize its non-iterative nature.

## 4.1 Two Datasets

We consider $d = 2$, and two cases. In the first case, we generate the signals $Z_i$ so that the individual components in JIVE decomposition are orthogonal, and JIVE model coincides with SLIDE model. We use this case to compare the performance of the methods with varying scale and size of the two datasets. In the second case, we generate the signals $Z_i$ so that the individual components in JIVE are not orthogonal, and hence JIVE model differs from SLIDE (see Appendix A). We note that in **r.jive** package, orthogonality is enforced by default (orthIndiv= TRUE). Therefore, to make the comparison fair, we use the default option (orthIndiv= TRUE) for the first case, and set it to FALSE for the second case.

### 4.1.1 Case 1: JIVE model and SLIDE Model Coincide

We generate $Z_1 \in \mathbb{R}^{n \times p_1}$ and $Z_2 \in \mathbb{R}^{n \times p_2}$ with the rank of shared structure $r_0 = 2$, and individual ranks $r_1 = r_2 = 2$ according to the following model:

$$Z_1 = c_1(U_1 D_0 W_{1,1}^\top + U_2 D_1 W_{1,2}^\top)$$
$$Z_2 = c_2(U_1 D_0 W_{2,1}^\top + U_3 D_2 W_{2,3}^\top),$$

where the scores $U_1 \in \mathbb{R}^{n \times 2}$, $U_2 \in \mathbb{R}^{n \times 2}$ and $U_3 \in \mathbb{R}^{n \times 2}$ are generated using uniform distribution on $[0, 1]$ with subsequent centering and orthonormalization of $U = [U_1 \, U_2 \, U_3]$. We set $D_0 = \text{diag}(1.5, 1.3)$, $D_1 = \text{diag}(1, 0.8)$, $D_2 = \text{diag}(1, 0.7)$, the loadings $W_{1,1}$, $W_{1,2}$, $W_{2,1}$ and $W_{2,3}$ are generated using uniform distribution on $[0, 1]$ with subsequent orthonormalization of $W = \begin{pmatrix} W_{1,1} & W_{1,2} & 0 \\ W_{2,1} & 0 & W_{2,3} \end{pmatrix}$, $c_1$, $c_2$ are nonzero constants. The orthonormalization of $W$ guarantees the uniqueness of SLIDE decomposition in accordance with Theorem 1, and subsequently ensures that it coincides with JIVE.

We further consider three combinations of $(p_1, p_2)$ and $(c_1, c_2)$:

1. $p_1 = p_2 = 25$, $c_1 = c_2 = 1$ (equal number of measurements and same scale)

2. $p_1 = p_2 = 25$, $c_1 = 0.5$, $c_2 = 1.5$ (equal number of measurements, but different scale)

3. $p_1 = 25$, $p_2 = 150$, $c_1 = c_2 = 1$ (different number of measurements, but same scale)

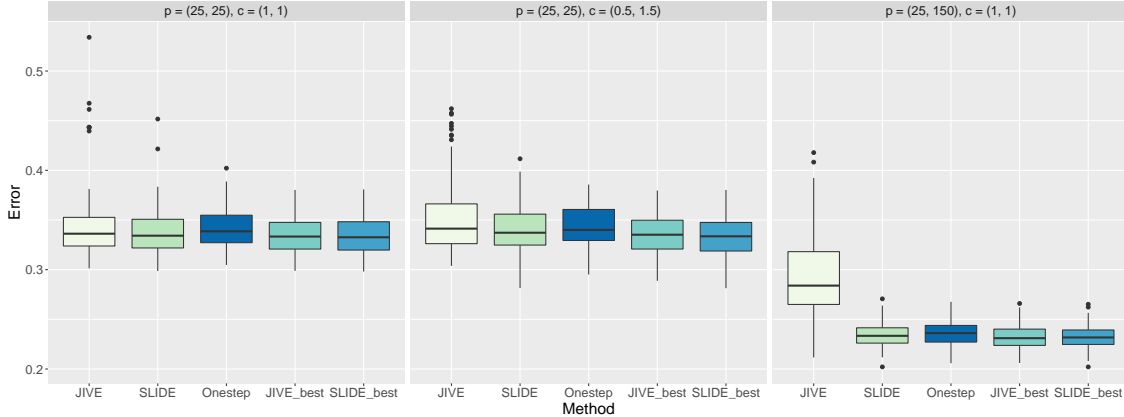

Figure 2: Two matched datasets, JIVE model coincides with SLIDE model. Displayed are scaled squared Frobenius norm errors (11) over 100 replications.

The scaled Frobenius norm error (11) over 100 replications is displayed in Figure 2. We note that the results with different scale are qualitatively similar to the results with the same scale, confirming that both the chosen error metric and the methods are scale-invariant. As expected, the error is smaller for the third scenario due to the larger value of $p_2$, which leads to the stronger signal for the same ranks and same initialization of noise variance (11).

In scenarios 1 and 2, SLIDE and JIVE have comparable performance, and both perform slightly worse than their "best" versions. This is not surprising since both JIVE and SLIDE sometimes estimate the ranks incorrectly as shown in Figure 3. Unexpected to us, SLIDE significantly outperforms JIVE in the 3rd scenario as the latter consistently overestimates the individual rank of 2nd dataset (with larger $p$). This suggests that permutation-based approach for rank selection in JIVE may lead to overfitting when one dataset has much larger number of measurements compared to the other. This conclusion is supported by the observed difference between JIVE and JIVE_best in Figure 2. In contrast, the SLIDE error is almost identical to SLIDE_best across all three scenarios. Moreover, Figure 3 reveals that the structure selected by SLIDE overwhelmingly coincides with true structure.

#### 4.1.2 Case 2 : SLIDE Model Differs from JIVE

We generate rank one signal matrices $Z_1 \in \mathbb{R}^{n \times p_1}$ and $Z_2 \in \mathbb{R}^{n \times p_2}$ as

$$Z_1 = u_1 v_1^\top, \quad Z_2 = u_1 v_2^\top,$$

where the scores $u_1 \in \mathbb{R}^{n \times 1}$ and $u_2 \in \mathbb{R}^{n \times 1}$ are such that $u_1^\top u_2 = c = 0.8$. Specifically, we first generate $\widetilde{U} = [\widetilde{u}_1 \widetilde{u}_2]$ using uniform distribution on $[0, 1]$ with subsequent centering and orthonormalization of $\widetilde{U}$, and then set $u_1 = \widetilde{u}_1$, $u_2 = (\alpha \widetilde{u}_1 + (1-\alpha)\widetilde{u}_2)/\sqrt{\alpha^2 + (1-\alpha)^2}$, where $\alpha = c/(c + \sqrt{1-c^2})$. The loading vectors $v_1, v_2$ are generated using uniform distribution on $[0, 1]$ with subsequent normalization so that $\|v_1\|_2 = \|v_2\|_2 = 1$. This model coincides with the toy example described in Section 2.1. Recall that according to JIVE model, $u_1$ and $u_2$ should be treated as individual scores, and rank of shared structure is zero. SLIDE model, on the other hand, should find one shared component and one individual component since $u_1^\top u_2 \neq 0$.

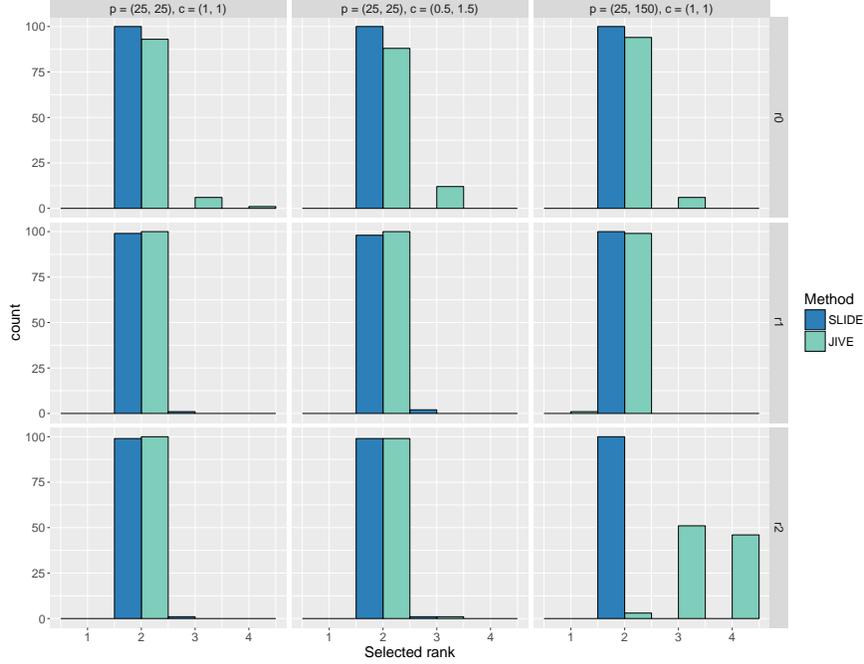

Figure 3: Two matched datasets, JIVE model coincides with SLIDE model. Displayed are counts for selected values for the rank of shared structure ($r_0$) and the ranks of individual structures ($r_1$, $r_2$) out of 100 replications. The true rank values are $r_0 = r_1 = r_2 = 2$.

The scaled Frobenius norm error (11) over 100 replications is displayed on the left panel in Figure 4, and the selected ranks for each type of structure are displayed on the right panel of Figure 4. We note that the absence of orthogonality between individual scores in JIVE leads to incorrect rank identification (JIVE consistently identifies rank 3 as the total rank for concatenated dataset), and subsequently increased Frobenius norm error. In contrast, SLIDE correctly identifies total rank of concatenated dataset to be 2 and overwhelmingly selects rank 1 for shared structure. The individual structure is either present in the 1st dataset or in the 2nd dataset, but never in both, which is consistent with identifiability requirements and discussion of Section 2.1. We note that applying JIVE with given true ranks leads to the smallest error out of all methods, confirming that the reason for JIVE's unsatisfactory performance is the structure misidentification.

## 4.2 Three Datasets

We generate $Z_1 \in \mathbb{R}^{n \times p_1}$, $Z_2 \in \mathbb{R}^{n \times p_2}$ and $Z_3 \in \mathbb{R}^{n \times p_3}$ with $p_1 = p_2 = p_3 = 100$ according to the following model:

$$\begin{aligned}
Z_1 &= U_0 D_0 W_{1,0}^\top + U_{12} D_{12} W_{1,12} + U_{13} D_{13} W_{1,13} && + U_1 D_1 W_{1,1}^\top \\
Z_2 &= U_0 D_0 W_{2,0}^\top + U_{12} D_{12} W_{2,12} && + U_{23} D_{23} W_{2,23} + U_2 D_2 W_{2,2}^\top \\
Z_3 &= U_0 D_0 W_{3,0}^\top && + U_{13} D_{13} W_{3,13} + U_{23} D_{23} W_{3,23} + U_3 D_3 W_{3,3}^\top,
\end{aligned}$$

with the following ranks for shared, partially-shared, and individual structures: $r_0 = r_1 = r_2 = r_2 = r_{12} = r_{13} = r_{23} = 2$. The scores $U_0, U_{12}, U_{13}, U_{23}, U_1, U_2, U_3 \in \mathbb{R}^{n \times 2}$ are generated using uniform dis-tribution on $[0, 1]$ with subsequent centering and orthonormalization of $U = [U_0\, U_{12}\, U_{13}\, U_{23}\, U_1\, U_2\, U_3] \in \mathbb{R}^{n \times 14}$, $D_0 = \text{diag}(1.5, 1.3)$, $D_{12} = \text{diag}(1, 0.8)$, $D_{13} = \text{diag}(1, 0.7)$, $D_{23} = \text{diag}(1, 0.5)$, $D_1 = \text{diag}(1.2, 0.5)$, $D_2 = \text{diag}(0.9, 0.8)$, $D_3 = \text{diag}(0.5, 0.4)$. The loadings $W_{d,\cdot}$ are generated using uniform distribution on $[0, 1]$ with subsequent orthonormalization of

$$W = \begin{pmatrix} W_{1,0} & W_{1,12} & W_{1,13} & 0 & W_{1,1} & 0 & 0 \\ W_{2,0} & W_{2,12} & 0 & W_{2,23} & 0 & W_{2,2} & 0 \\ W_{3,0} & 0 & W_{3,13} & W_{3,23} & 0 & 0 & W_{3,3} \end{pmatrix}.$$

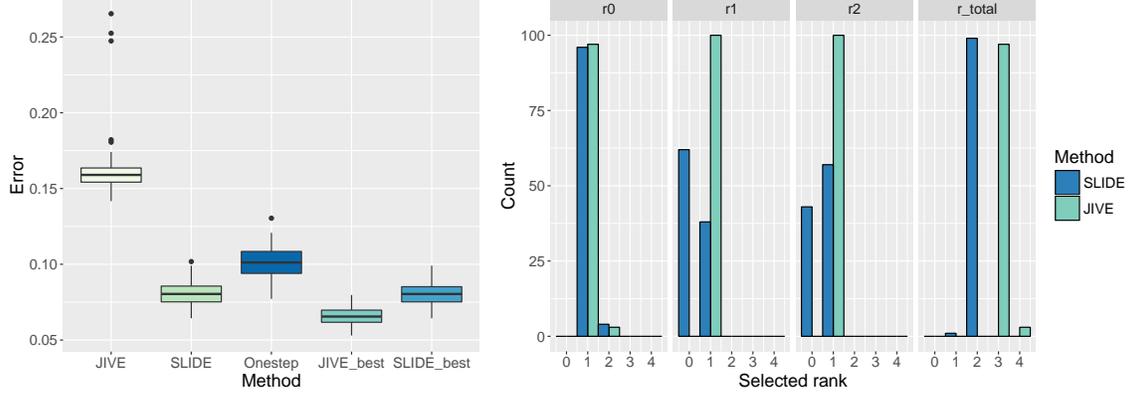

Figure 4: Two matched datasets, JIVE model differs from SLIDE model. Left panel: Scaled squared Frobenius norm errors (11) over 100 replications. Right panel: Selected values for the rank of shared structure ($r_0$) and the ranks of individual structures ($r_1$, $r_2$) out of 100 replications. Both signal matrices have rank one with non-orthogonal column spaces. JIVE true ranks are $r_0 = 0$, $r_1 = r_2 = 1$. SLIDE true ranks are $r_0 = 1$, and $r_1 + r_2 = 1$. Both cases $r_1 = 1, r_2 = 0$ and $r_1 = 0, r_2 = 1$ satisfy the SLIDE model.

The orthogonality of $W$ ensures that the definition of shared structure coincides between JIVE and SLIDE. The partially-shared structures, however, should be treated as individual in JIVE model due to zero intersection of the three corresponding column spaces.

The scaled Frobenius norm error (11) over 100 replications is displayed on the left panel in Figure 5. While JIVE outperforms one-step approach on true ranks, it performs significantly worse than the SLIDE method. The right panel of Figure 5 indicates that the selected structure coincided with the true structure in the majority of replications. Since JIVE model doesn't take into account partially-shared structures, we compare the total estimated rank for concatenated dataset $X$ as well as total estimated individual ranks for each dataset in Figure 6. We note that these ranks are invariant to the definition of shared, partially-shared and individual structures. We observe that JIVE consistently overestimates both the total rank (true rank is 14), and individual ranks (true rank is 8 for each), confirming that rank selection scheme is sensitive to the presence of partially-shared structures. In contrast, SLIDE correctly identifies the ranks in the majority of replications.

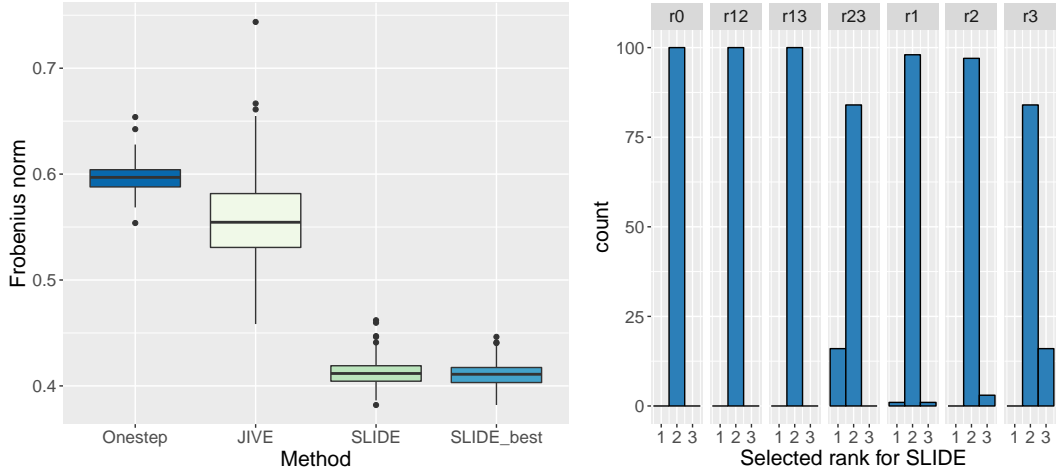

Figure 5: Three matched datasets. Left panel: Scaled squared Frobenius norm errors (11) over 100 replications. Right panel: Ranks selected by BCV for SLIDE model, the true ranks are all equal to two.

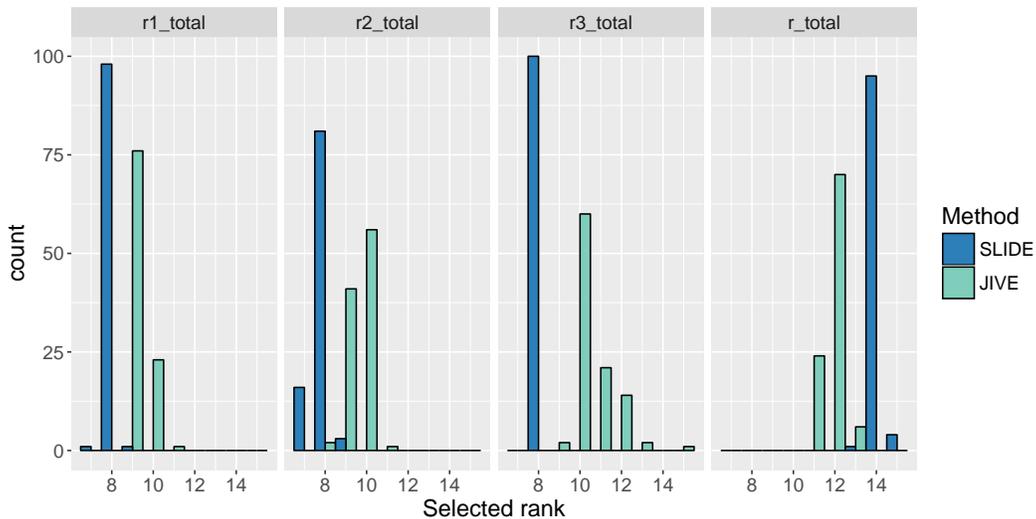

Figure 6: Three matched datasets. Selected total ranks for individual datasets ($ri\_total$, $i = 1, 2, 3$) and concatenated dataset ($r\_total$) for SLIDE and JIVE over 100 replications. The true ranks are 8 for each individual dataset, and 14 for the concatenated dataset.

## 5 Application to TCGA BRCA Data

We apply the proposed SLIDE method to multi-view data on breast cancer from TCGA. In particular, we consider 4 different data sources: gene expression (GE), DNA methylation (ME), miRNA expression (miRNA), and reverse phase protein array (RPPA). The data are publicly available at https://tcga-data.nci.nih.gov/docs/publications/brca_2012/. We follow the same procedure as in Lock and Dunson (2013) to preprocess the multi-view data. More specifically, we first restrict our scope to the 348 common samples with measurements in all data types. Next, we im-pute missing values in GE with the K-nearest neighbor algorithm ($K = 10$), and filter the imputed data with column-wise standard deviation threshold 1.5. We apply square root transformation to ME, and log transformation to miRNA after removing variables with zeros in more than half of the samples. Finally, all individual data matrices are column-centered and scaled to have unit Frobenius norms. As a result, we obtain $X_1 : 348 \times 645$ for GE, $X_2 : 348 \times 574$ for ME, $X_3 : 348 \times 423$ for miRNA, and $X_4 : 348 \times 171$ for RPPA. In addition, we also have matched demographics, breast cancer subtypes, and survival information on the 348 subjects. Below we compare SLIDE and JIVE in terms of model fitting, subtype classification, and clustering and survival analysis.

|       | SLIDE         | JIVE          |
| ----- | ------------- | ------------- |
| GE    | 14 (47.89%)   | 34 (61.72%)   |
| ME    | 10 (42.56%)   | 31 (47.87%)   |
| miRNA | 19 (57.50%)   | 30 (48.20%)   |
| RPPA  | 20 (63.55%)   | 22 (52.04%)   |
| Total | 50 (52.88%)   | 108 (52.46%)  |

Table 1: TCGA-BRCA rank estimation and variation explained. The table shows the ranks for different data sets estimated by different methods, as well as the percentage of variation explained by the corresponding low-rank structure (in parenthesis). For example, the total rank for GE estimated by SLIDE is equal to the globally shared rank (i.e., 3) plus the partially-shared rank involving GE (i.e., 3) plus its individual rank (i.e., 8). The percentage of variation explained is calculated by the squared Frobenius norm of the corresponding estimated low-rank structure divided by the squared Frobenius norm of data.

We first apply both methods to determine ranks of the underlying structures. JIVE estimates the total rank to be 108, with the globally shared rank being 3, and individual ranks being 31 (GE), 28 (ME), 27 (miRNA), and 19 (RPPA). In comparison, SLIDE estimates the total rank to be 50, much smaller than the JIVE estimate. The globally shared rank from SLIDE is 3, and the partially shared rank is 3: one for (GE, ME, miRNA), one for (GE, ME), and one for (GE, miRNA). The individual ranks are 8 (GE), 5 (ME), 14 (miRNA), and 17 (RPPA). In Table 1, we summarize the ranks and variation explained by both methods in different data sets. The ranks estimated by SLIDE are uniformly smaller than those by JIVE, but the percentages of variation explained are comparable. Surprisingly, the total variation explained by the 50 ranks in SLIDE is even higher than that in JIVE, possibly because the estimation procedure of JIVE is suboptimal compared to the proposed Algorithm 2 with given structures.

Next, we evaluate the subtype classification performance based on the shared score vectors from JIVE and SLIDE respectively. For fair comparison, we only focus on the 3 globally shared score vectors from SLIDE. Figure 7 shows scatter plots of the shared scores estimated from JIVE (left) and SLIDE (right), color-coded by predefined breast cancer subtypes (basal (66), HER2 (42), LumA (154), LumB (81), normal-like (5)). The SLIDE scatter plots provide better separations among different subtypes. To quantify the classification performance, we calculate the SWISS score (Cabanski et al., 2010) for each method, which characterizes subtype distinctions using standardized within-subtype sum of squares. Smaller values indicate better distinctions of different subtypes. A more detailed description can be found in Cabanski et al. (2010). The SWISS score is 0.6288 for JIVE, and 0.5061 for SLIDE. Namely, the shared scores of SLIDE indeed provide superior subtype classification performance over JIVE.

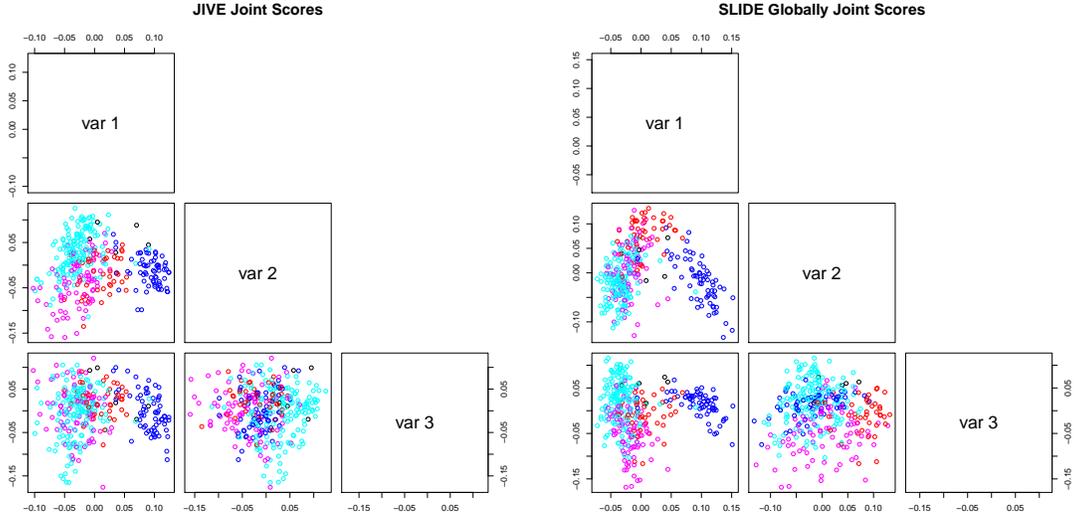

Figure 7: TCGA-BRCA subtype classification by JIVE and SLIDE. The left panels are scatter plots of JIVE shared scores and the right panels are scatter plots of SLIDE globally shared scores. The scatter plots are color-coded by predefined breast cancer subtypes: basal (blue), HER2 (red), LumA (cyan), LumB (magenta), and normal-like (black).

In addition, we also carry out exploratory clustering analysis on subjects. To reduce heterogeneity, we first remove 5 subjects with the normal-like subtype. Then we conduct hierarchical clustering based on the 3 shared scores from JIVE, and the 6 (both globally and partially) shared scores from SLIDE, respectively. We use the Euclidean distance metric and the Ward's minimum variance method in the hierarchical clustering. To evaluate the clustering performance, we explicitly identify 4 subgroups from each method, and (i) compare subjects' memberships with their predefined subtypes; (ii) compare the Kaplan-Meier curves corresponding to different subgroups using log-rank test. Table 2 presents the membership relations between the predefined subtypes and the subgroups induced from different methods. The SLIDE subgroups seem to have more concordance with existing breast cancer subtypes than the JIVE subgroups. In particular, almost all Basal subjects belong to the second SLIDE group, and almost all Her2 subjects belong to the third SLIDE group. LumA and LumB subjects are spread across the first, third and fourth groups, with most LumA in the first group (over 61%) and most LumB in the fourth group (over 40%). We further compare the Kaplan-Meier curves for different subgroups induced by JIVE and SLIDE in Figure 8. The log-rank test on the differences between different Kaplan-Meier curves from SLIDE provides a p-value of 0.029. Namely, different subgroups have significantly distinct survival behaviors. The result is not significant (p-value 0.375) for the subgroups induced by JIVE. Consequently, the shared structure identified by SLIDE proves more insightful and useful in defining new disease subtypes than JIVE.

|     |       | JIVE Group1 | JIVE Group2 | JIVE Group3 | JIVE Group4 |
|-----|-------|-------------|-------------|-------------|-------------|
|     | Basal | 3           | 63          | 0           | 0           |
| (a) | HER2  | 17          | 9           | 15          | 1           |
|     | LumA  | 102         | 2           | 27          | 23          |
|     | LumB  | 18          | 0           | 52          | 11          |

|     |       | SLIDE Group1 | SLIDE Group2 | SLIDE Group3 | SLIDE Group4 |
|-----|-------|--------------|--------------|--------------|--------------|
|     | Basal | 1            | 61           | 4            | 0            |
| (b) | HER2  | 4            | 0            | 38           | 0            |
|     | LumA  | 94           | 3            | 21           | 36           |
|     | LumB  | 16           | 1            | 31           | 33           |

Table 2: TCGA-BRCA hierarchical clustering analysis based on 3 shared scores from JIVE, and 6 globally and partially-shared scores from SLIDE. (a) Membership relations between predefined subtypes and exploratory subgroups induced from JIVE; (b) membership relations between predefined subtypes and exploratory subgroups induced from SLIDE.

## 6 Discussion

In this work, we propose a new structured decomposition for multi-view data called SLIDE, which allows to directly model shared, partially-shared and individual components. Our empirical studies demonstrate the superiority of SLIDE over existing methods in terms of rank selection, and underlying low-rank structure recovery. The computational efficiency is achieved by performing a significant reduction in Step 1 of the number of structures to be considered (see Figure 1). We note that the reduction is necessary for any structure selection approach to be executable within polynomial time (see Proposition 1). Our empirical studies indicate that using the support from the path of solutions for the penalized matrix factorization problem (8) leads to the reduced set that contains the true structure in the majority of the replications. It it of great interest to support these empirical findings with theoretical analysis, and in particular, understand the required assumptions on the signal so that the correct structure is contained within the path of problem (8) with high probability. We expect this to be a very challenging problem due to non-convexity of problem (8), which will require further investigation.

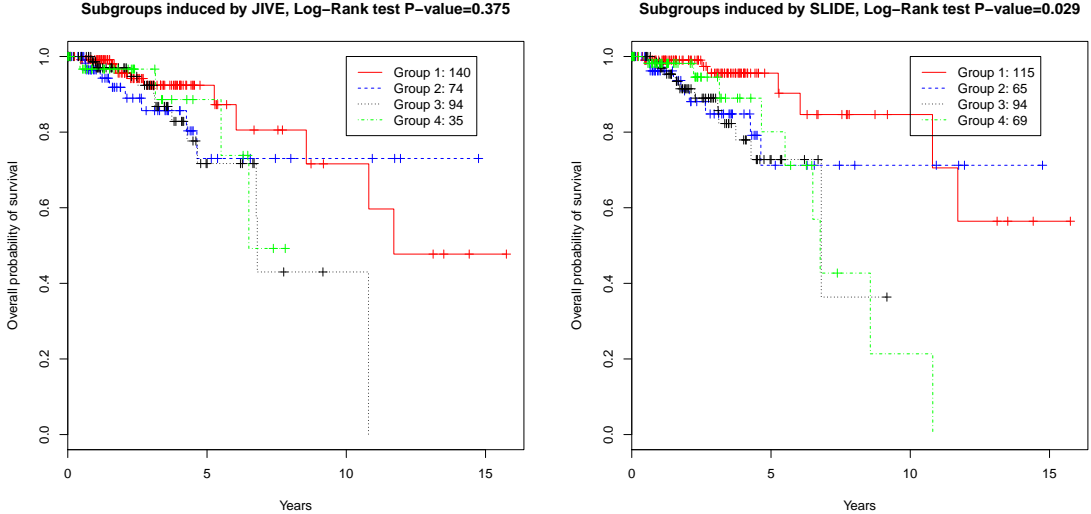

Figure 8: TCGA-BRCA Kaplan-Meier curves for different subgroups induced by JIVE (left) and SLIDE (right).

## A  Discussion of SLIDE Model Identifiability and Comparison with JIVE in Case of Two Datasets

In this section, we further discuss identifiability conditions for the SLIDE model. We use JIVE (Lock et al., 2013) identifiability conditions as a reference, and to further illustrate the similarities and differences between the two approaches. We limit the discussion to the case of two datasets as the JIVE model does not allow for partially-shared components.

The JIVE considers the following additive noise decomposition

$$X_i = J_i + A_i + E_i, \quad i = 1, 2. \tag{A.1}$$

Here $J_i \in \mathbb{R}^{n \times p_i}$ represent the shared structures across two datasets with $\text{rank}(J_i) = r_0$, $A_i \in \mathbb{R}^{n \times p_i}$ represent individual structures of each dataset $i$ with $\text{rank}(A_i) = r_i$, and $E_i \in \mathbb{R}^{n \times p_i}$ are error matrices with independent entries of mean zero. We let $Z_i = J_i + A_i$ denote the low rank representation of each $X_i$.

**Proposition 4.** *(Lock et al., 2013; Feng et al., 2017) Given a set of matrices $\{Z_i, i = 1, 2\}$, there are unique sets of matrices $\{J_i, i = 1, 2\}$ and $\{A_i, i = 1, 2\}$ such that*

1. $Z_i = J_i + A_i$ for $i = 1, 2$
2. $\text{row}(J_i) \cap \text{row}(A_i) = \{0\}$ for $i = 1, 2$
3. $\text{col}(J_i) \perp \text{col}(A_i)$ for $i = 1, 2$
4. $\text{col}(J_i) = \text{col}(J)$ for $i = 1, 2$, where $J = [J_1 \ J_2]$
5. $\text{col}(A_1) \cap col(A_2) = \{0\}$.

**Remark 3.** *Condition 2 on the row spaces is equivalent to rank condition in Lock et al. (2013), and $\operatorname{col}(J) \subset \operatorname{col}(Z_i), i = 1, 2$ condition in Feng et al. (2017). It is necessary for the uniqueness of JIVE decomposition as together with condition 3 it ensures that $\operatorname{col}(J_i + A_i) = \operatorname{col}(J_i) \cup \operatorname{col}(A_i)$.*

Thus, the signal part of JIVE model (A.1) has unique representation under conditions 2-5. The matrices $J_i$ are shared due to the common column space, and matrices $A_i$ are individual due to zero intersection of their respective column spaces. The individual $A_i$, however, are not required to be orthogonal. We first show that the matrices $A_i$ can be further decomposed into two parts, one of which is the orthogonal signal.

**Proposition 5.** *Given a set of matrices $\{Z_i, i = 1, 2\}$ with the unique decomposition $Z_i = J_i + A_i$ from Proposition 4, there exist a unique set of matrices $\{N_i, i = 1, 2\}$ and $\{O_i, i = 1, 2\}$ such that*

1. $A_i = N_i + O_i$ for $i = 1, 2$

2. $\operatorname{row}(N_i) \cap \operatorname{row}(O_i) = \{0\}$ for $i = 1, 2$

3. $\operatorname{col}(N_i) \perp \operatorname{col}(O_j)$ for all $i, j = 1, 2, i \neq j$

4. $\operatorname{col}(O_1) \perp \operatorname{col}(O_2)$

5. $\operatorname{col}(N_1) \cap \operatorname{col}(N_2) = \{0\}$ and the maximal principle angle between $\operatorname{col}(N_1)$ and $\operatorname{col}(N_2)$ is strictly smaller than $\pi/2$.

Proposition 5 decomposes each individual $A_i$ of JIVE into two parts. One part, $O_i$, is completely orthogonal to $A_j$ for $j \neq i$, and hence is truly unique to the dataset $i$. On the other hand, $N_1$ and $N_2$ have non-zero and non-orthogonal principal angles, meaning that $\operatorname{rank}(N_1) = \operatorname{rank}(N_2) = \operatorname{rank}(N_1^\top N_2) = r_N$. In the terminology of canonical correlation analysis, this implies the existence of $r_N$ orthogonal pairs of canonical variables $\{w_{1j}, w_{2j}\}$, $j = 1, \ldots, r_N$, with non-zero correlations between $A_1 w_{1j}$ and $A_2 w_{2j}$. Since $N_i$ are within the individual structures $A_i$, the shared structures $J_i$ in JIVE do not fully capture all associations present between the datasets.

We further show that it is possible to decompose $N_i$ into shared and orthogonal components at the expense of the uniqueness of decomposition.

**Proposition 6.** *Given a set of matrices $\{Z_1, Z_2\}$, consider matrices $\{N_i, i = 1, 2\}$ from Propositions 5.*

1. *There exist matrices $C_i$, $I_i$ such that*

    (a) $N_i = C_i + I_i$ for $i = 1, 2$

    (b) $\operatorname{col}(C_i) = \operatorname{col}([C_1 \ C_2])$ for $i = 1, 2$

    (c) $\operatorname{col}(C_i) \perp \operatorname{col}(I_i)$ for $i = 1, 2$

    (d) $\operatorname{col}(I_1) \perp \operatorname{col}(I_2)$

    (e) $\operatorname{row}([C_1 \ C_2]) \cap \operatorname{row}([I_1 \ \mathbf{0}]) = \{0\}$, $\operatorname{row}([C_1 \ C_2]) \cap \operatorname{row}([\mathbf{0} \ I_2]) = \{0\}$, where $\mathbf{0}$ is the zero matrix of compatible size.

2. *Matrix $C_1^\top C_2$ is unique, and $\operatorname{rank}(C_i) = \operatorname{rank}([C_1 \ C_2]) = \dim(\operatorname{col}(I_1)) + \dim(\operatorname{col}(I_2)) = \operatorname{rank}(N_i) = \operatorname{rank}(N_1^\top N_2)$.*

Combining the above results leads to modified decomposition of signal matrices $Z_1, Z_2$ that has orthogonal individual structures.

**Corollary 1.** *Given a set of matrices $\{Z_i, i = 1, 2\}$,*

1. *there exist matrices $\{J'_i, i = 1, 2\}$, $\{A'_i, i = 1.2\}$ such that*

    (a) $Z_i = J'_i + A'_i$ for $i = 1, 2$
    (b) $\operatorname{col}(J'_i) \perp \operatorname{col}(A'_i)$ for $i = 1, 2$
    (c) $\operatorname{col}(J'_i) = \operatorname{col}([J'_1 \ J'_2])$ for $i = 1, 2$
    (d) $\operatorname{col}(A'_1) \perp \operatorname{col}(A'_2)$
    (e) $\operatorname{row}(J') \cap \operatorname{row}([A'_1 \ \mathbf{0}]) = \{0\}$, $\operatorname{row}(J') \cap \operatorname{row}([\mathbf{0} \ A'_2]) = \{0\}$, where $\mathbf{0}$ is the zero matrix of compatible size.

2. *If matrices $N_i$ from Proposition 5 are zero, then $J'_i$ and $A'_i$ are unique, and correspond to $J_i$, $A_i$ from Proposition 4.*

3. *If $N_i \neq 0$, then $J'_i = J_i + C_i$, $A'_i = I_i + O_i$, where $J_i$, $O_i$ and $C_i + I_i = N_i$ are unique, and $C_i$, $I_i$ are any matrices satisfying Proposition 6.*

Comparison with Theorem 1 reveals that SLIDE model corresponds to the decomposition of Corollary 1. Let $r = \operatorname{rank}(Z)$, and consider $S \in \{0, 1\}^{2 \times r}$ with non-zero columns such that SLIDE decomposition (5) holds. Let $F \subseteq \{1, .., r\}$ be the index set of shared scores ($s_i = (1, 1)$ for $i \in F$), $F_1 \subseteq \{1, .., r\}$ be the index set of individual scores for matrix $Z_1$ ($s_i = (1, 0)$ for $i \in F_1$), and $F_2 \subseteq \{1, .., r\}$ be the index set of individual scores for matrix $Z_2$ ($s_i = (0, 1)$ for $i \in F_2$). Then it follows that

$$Z_1 = U_F V_{1,F}^\top + U_{F_1} V_{1,F_1}^\top = J'_1 + A'_1$$
$$Z_2 = U_F V_{2,F}^\top + U_{F_2} V_{2,F_2}^\top = J'_2 + A'_2,$$

where $J'_1, A'_1, J'_2, A'_2$ satisfy the requirements of Corollary 1.

In case $N_i$ from Proposition 5 are zero, the decomposition is unique and SLIDE coincides with JIVE. In case $N_i \neq 0$, the decomposition is not unique. The non-uniqueness is the price to pay for the orthogonality requirement, which allows correct shared component identification in a sense of canonical correlation analysis. To clearly see the latter, consider the cross-covariance structure implied by the two models. Under the JIVE model from Proposition 4,

$$Z_1^\top Z_2 = J_1^\top J_2 + J_1^\top A_2 + A_1^\top J_2 + A_1^\top A_2 = J_1^\top J_2 + A_1^\top A_2,$$

which includes both the shared terms and the individual terms unless $A_1 \perp A_2$. Under the model from Corollary 1,

$$Z_1^\top Z_2 = {J'_1}^\top J'_2 + {J'_1}^\top A'_2 + {A'_1}^\top J'_2 + {A'_1}^\top A'_2 = {J'_1}^\top J'_2,$$

which only includes the shared terms as desired.

# B  Technical Proofs

In this section we provide the proofs of the technical results of the paper.

## B.1 Proof of Theorem 1

*Proof.* **Part 1.** Consider the signal matrices $Z_1, \ldots, Z_d$ corresponding to $d$ matched datasets $X_1, \ldots, X_d$ such that $X_i = Z_i + E_i$ with noise perturbation matrices $E_i$, and let $Z = [Z_1 \ldots Z_d]$. The SLIDE decomposition with specified requirements can be constructed sequentially given fixed $Z$ by first separating individual scores, then partially-shared scores, and finally globally shared scores. Without loss of generality, we illustrate how to construct such decomposition for $d = 3$ case, the extension to $d > 3$ is straightforward by adding partially shared structures of higher order. We note that this construction is purely of theoretical value as it shows the existence of decomposition. In practice, signal $Z$ is always observed with noise, so direct construction is not possible.

Let $R_1, R_2, R_3$ be equal to $Z_1, Z_2, Z_3$ correspondingly. Matrices $R_1, R_2, R_3$ will be updated throughout the steps of the construction, whereas signal matrices $Z_1, Z_2, Z_3$ will remain fixed. We use $P_A$ to denote the projection matrix on the column space of $A$, and $P_{A^\perp}$ onto the orthogonal space. Then the following steps lead to SLIDE decomposition that satisfies the requirements of proposition:

1. Construction of individual scores.

    For datasets $i = 1, 2, 3$

    - set $U^{(i)}$, the individual components for dataset $i$, to be the orthonormal basis vectors for $\mathrm{col}(P_{R_k^\perp} P_{R_j^\perp} R_i)$, where $j, k, i$ are distinct
    - set $V^{(i)} = (P_{R_k^\perp} P_{R_j^\perp} R_i)^\top U^{(i)}$ to be the loading vectors corresponding to $U^{(i)}$
    - update $R_i = (I - P_{R_k^\perp} P_{R_j^\perp}) R_i$

2. Construction of partially shared scores.

    For all pairs $i, j \in \{1, 2, 3\}$ with $i \neq j$

    - set $U^{(ij)}$, the partially-shared components between datasets $i$ and $j$, to be the orthonormal basis vectors for $\mathrm{col}(P_{R_k^\perp} [R_i\ R_j])$, where $k \neq i$, $k \neq j$
    - set $V^{(ij)} = (P_{R_k^\perp} [R_i\ R_j])^\top U^{(ij)}$ to be the loading vectors corresponding to $U^{(ij)}$
    - update $[R_i\ R_j] = (I - P_{R_k^\perp})[R_i\ R_j]$

3. Construction of shared scores.

    - set $U^{(123)}$, the globally shared components, to be the orthonormal basis vectors for $\mathrm{col}([R_1\ R_2\ R_3])$
    - set $V^{(123)} = ([R_1\ R_2\ R_3])^\top U^{(123)}$ to be the loading vectors corresponding to $U^{(123)}$

By construction, it holds that $Z = UV^\top$ with $U = [U^{(123)}\ U^{(12)}\ U^{(13)}\ U^{(31)}\ U^{(1)}\ U^{(2)}\ U^{(3)}]$ satisfying $U^\top U = I$, and $V$ formed from $V^{(l)}$ above by appending zeroes to match the dimensions (that is $V^{(1)}$ becomes $\begin{bmatrix} V^{(1)\top} & 0 & 0 \end{bmatrix}^\top$). Moreover, by construction, each $V^{(l)}$ has linearly independent columns. Finally, we note that changing the order of datasets in part 1, and the order of pairs in part 2, may lead to a different SLIDE decomposition that will also satisfy the requirements. However, in certain cases the order is irrelevant leading to unique decomposition. Such cases are formalized in part 2.

**Part 2.** Consider the signal part of SLIDE decomposition (5), $Z = UV^\top$. First, we will show that the linear independence of nonzero columns in dataset-specific blocks $V_i$ leads to uniqueness of $S$ according to Definition 1. Secondly, we will show that the uniqueness of $S$ together with orthogonality requirement implies the uniqueness of $U$ and $V$. Throughout the proof, we let $\mathcal{B}_d$ be the set of distinct nonzero binary vectors $b_k$ of length $d$, $k = 1, \ldots, 2^d - 1$. For any $S$, we let $r_k$ denote the number of columns in $S$ with sparsity pattern according to $b_k \in \mathcal{B}_d$. Then $S$ is uniquely determined by $r_1, \ldots, r_{2^d-1}$, therefore it is sufficient to show uniqueness of $r_1, \ldots, r_{2^d-1}$.

We will first show that for any subset $G \subseteq \{1, \ldots, d\}$, the nonzero columns of $V_G$ are linearly independent, where $V_G$ is formed by appending dataset-specific blocks $V_i$, $i \in G$, row-wise. Let $g$ be the cardinality of $G$ with $g = d$ when $V_G = V$, and $g = 1$ when $V_G = V_i$ for some $i = 1, \ldots, d$. Then for $g = 1$, the claim follows from linear independence of $V_i$, so it remains to consider case $g > 1$. Let $\mathcal{B}_g$ be the set of distinct binary vectors of length $g$, and let $V_{G,b_k}$ be the subset of columns of $V_G$ with sparsity pattern according to $b_k \in \mathcal{B}_g$. Then by construction, $\mathrm{col}(V_{G,b_k}) \cap \mathrm{col}(V_{G,b_j})$ when $b_k \neq b_j$, so it remains to show that the nonzero columns of $V_{G,b_k}$ are linearly independent for each $b_k$. Assume this is not the case for some $b_k \in \mathcal{B}_g$, and that there are $l$ corresponding nonzero columns. Then there exists nonzero vector $c = (c_1, \ldots, c_l)$ such that $\sum_{j=1}^{l} c_j \{V_{G,b_k}\}_j = 0$. It follows that for any dataset $i \in G$ such that $b_{ki} \neq 0$, $\sum_{j=1}^{l} c_j \{V_{i,b_k}\}_j = 0$, which contradicts linear independence of nonzero columns in $V_i$. Hence, the nonzero columns of $V_G$ are linearly independent for each $G \subseteq \{1, \ldots, d\}$.

We will now show that linear independence of nonzero columns of $V_G$ implies uniqueness of $S$ in SLIDE decomposition. For any $G \subseteq \{1, \ldots, d\}$, consider the signal matrix $Z_G$ formed by concatenating signal matrices $Z_i$, $i \in G$. Then from SLIDE decomposition

$$Z_G = UV_G^\top = \sum_{b_k \in \mathcal{B}_d} U_{b_k} V_{G,b_k}^\top = \sum_{b_k \in \mathcal{B}_d, b_{kG} \neq 0} U_{b_k} V_{G,b_k}^\top,$$

where $b_{kG}$ is the subvector of $b_k$ corresponding to elements in $G$, and $U_{b_k}$ is the submatrix of $U$ with columns corresponding to the sparsity pattern $b_k$. The last equality above eliminates the zero columns in $V_G$. Using orthonormality of the columns of $U$ together with linear independence of nonzero columns of $V_G$ leads to

$$\mathrm{rank}(Z_G) = \sum_{b_k \in \mathcal{B}_d, b_{kG} \neq 0} \mathrm{rank}(U_{b_k} V_{G,b_k}^\top) = \sum_{b_k \in \mathcal{B}_d, b_{kG} \neq 0} r_k,$$

where $r_k$ is the number of columns in $S$ with sparsity pattern according to $b_k$, and $\mathrm{rank}(Z_G)$ is unique for each $G$. Combining all distinct nonzero subsets $G \subseteq \{1, \ldots, d\}$ leads to $2^d - 1$ distinct linear equations of the form above with $2^d - 1$ unknowns ($r_1, \ldots, r_{2^d-1}$). It follows that the combination $r_1, \ldots, r_{2^d-1}$ that satisfies all equations must be unique, hence $S$ is unique.

Next, we show that orthogonality of the columns of $V$ corresponding to each $b_k \in \mathcal{B}_d$ ($V_{b_k}$) leads to uniqueness of $U$ and $V$. For any $b_k \in \mathcal{B}_d$, let $F_k \subseteq \{1, \ldots, d\}$ be the index set of non-zero elements in $b_k$, and $F_k^c = \{1, \ldots, d\} \setminus F_k$. Consider the projection matrix $P(b_k)$ defined as $P(b_k) = \prod_{j \in F_k^c} P_{Z_j^\perp} \prod_{i \in F_k} P_{Z_i}$. Since linear independence of non-zero columns in $V_i$ implies $\mathrm{col}(Z_i) = \mathrm{col}(U_j, s_{ij} \neq 0)$ and the columns of $U$ are orthonormal, it follows that

$$P(b_k) Z = \prod_{j \in F_k^c} P_{Z_j^\perp} \prod_{i \in F_k} P_{Z_i} Z = U_{b_k} V_{b_k}^\top.$$

Due to orthogonality of $V_{b_k}$, there exist $Q \in \mathbb{R}^{p \times r_k}$ with orthonormal columns and diagonal $D \in \mathbb{R}^{r_k \times r_k}$ with non-zero diagonal elements such that $V_{b_k} = QD$. Therefore,

$$\prod_{j \in F_k^c} P_{Z_j^\perp} \prod_{i \in F_k} P_{Z_i} Z = U_{b_k} D Q^\top,$$

where the right hand side is the singular value decomposition of $\prod_{j \in F_k^c} P_{Z_j^\perp} \prod_{i \in F_k} P_{Z_i} Z$. Since the signals $Z_i$ are all fixed, and the projection matrix is unique, it follows that $U_{b_k}$ and $V_{b_k}$ are unique for each $b_k \in \mathcal{B}_d$. Together with uniqueness of $S$, this implies the uniqueness of $U$ and $V$. $\square$

## B.2 Proof of Proposition 1

*Proof.* Note that each column of $S \in \{0,1\}^{d \times r}$ is equal to one of the $2^d$ distinct binary vectors $b_1, ..., b_{2^d}$ (including zero vector). Denote by $r_1, ..., r_{2^d}$ the number of columns that are equal to $b_1, ..., b_{2^d}$ correspondingly. Then $\sum_{i=1}^{2^d} r_i = r$, and using Definition 1, $S$ is uniquely determined by $r_1, ..., r_{2^d}$. Therefore, the number of distinct $S$ is the same as the number of distinct solutions to integer equation

$$\sum_{i=1}^{2^d} r_i = r, \quad r_i \geq 0.$$

This equation corresponds to the occupancy numbers problem, and has $\binom{r+2^d-1}{2^d-1}$ distinct solutions (Feller, 1968, Chapter 2.5). $\square$

## B.3 Proof of Proposition 2

*Proof.* When $V = 0$, the objective function in (8) is equal to $\frac{1}{2}\|X\|_F^2$, therefore any $\lambda$ for which $V = 0$ must satisfy

$$\sum_{i=1}^d \left[ \frac{1}{2} \text{Tr}(V_i^\top V_i) - \text{Tr}(V_i^\top X_i^\top U) + \lambda \sum_{j=1}^r \|V_{ij}\|_2 \right] \geq 0. \tag{B.2}$$

From (B.2), it it sufficient for $\lambda$ to satisfy for every $i \in \{1, ..., d\}$

$$\frac{1}{2} \text{Tr}(V_i^\top V_i) - \text{Tr}(V_i^\top X_i^\top U) + \lambda \sum_{j=1}^r \|V_{ij}\|_2 \geq 0,$$

or equivalently

$$\frac{1}{2} \text{Tr}(V_i^\top V_i) \geq \text{Tr}(V_i^\top X_i^\top U) - \lambda \sum_{j=1}^r \|V_{ij}\|_2.$$

Using Cauchy-Schwarz inequality, if $\lambda \geq \max_j \|X_i^\top U_j\|_2$, then

$$\text{Tr}(V_i^\top X_i^\top U) - \lambda \sum_{j=1}^r \|V_{ij}\|_2 = \sum_{j=1}^r V_{ij}^\top X_i^\top U_j - \lambda \sum_{j=1}^r \|V_{ij}\|_2$$

$$\leq \sum_{j=1}^r \|V_{ij}\|_2 \|X_i^\top U_j\|_2 - \lambda \sum_{j=1}^r \|V_{ij}\|_2$$

$$= \sum_{j=1}^r \|V_{ij}\|_2 \left( \|X_i^\top U_j\|_2 - \lambda \right)$$

$$\leq 0.$$

Since $U$ satisfies $U^\top U = I$, it follows that

$$\max_{i,j} \|X_i^\top U_j\|_2 \leq \max_{i,j} \sqrt{U_j^\top X_i X_i^\top U_j} \leq \max_i \sigma_{\max}(X_i)$$

Therefore, (B.2) holds for any $\lambda$ such that $\lambda \geq \max_i \sigma_{\max}(X_i)$, which concludes the proof. $\square$

## B.4 Proof of Proposition 3

*Proof.* Let $U^{(k)}$, $V^{(k)}$ be the solutions at step $k$. Then

$$U^{(k)} \in \operatorname*{argmin}_{U : U^\top U = I} f(U, V^{(k)}) \quad \text{and} \quad V^{(k)} = \operatorname*{argmin}_V f(U^{(k-1)}, V).$$

It follows that for each $k$

$$f(U^{(k-1)}, V^{(k-1)}) \geq f(U^{(k-1)}, V^{(k)}) \geq f(U^{(k)}, V^{(k)}),$$

hence the sequence $f^{(k)} = f(U^{(k)}, V^{(k)})$, $k = 1, 2, \ldots$, is non-increasing. Since $f(U, V) \geq 0$, the sequence $f^{(k)}$ must have a limiting point $f^*$ at some $k = k^*$, that is

$$f^* = f(U^{(k^*-1)}, V^{(k^*-1)}) = f(U^{(k^*-1)}, V^{(k^*)}) = f(U^{(k^*)}, V^{(k^*)}) = f(U^{(k^*)}, V^{(k^*+1)}).$$

It follows that

$$U^{(k^*)} \in \operatorname*{argmin}_U f(U, V^{(k^*)}) \quad \text{and} \quad V^{(k^*+1)} = \operatorname*{argmin}_V f(U^{k^*}, V).$$

Since $f(U^{(k^*)}, V^{(k^*)}) = f(U^{(k^*)}, V^{(k^*+1)})$, and the function $f(U, V)$ is strictly convex with respect to $V$ due to orthogonality of $U$, we must also have

$$V^{(k^*)} = \operatorname*{argmin}_V f(U^{k^*}, V),$$

which concludes the proof. $\square$

## B.5 Proof of Proposition 5

*Proof.* **Existence.** Let $U_1 \in \mathbb{R}^{n \times r_1}$ be an orthonormal basis for $A_1$, and $U_2 \in \mathbb{R}^{n \times r_2}$ be an orthonormal basis for $A_2$. Without loss of generality, let $r_1 \leq r_2$. Consider full singular value decomposition
$$U_1^\top U_2 = Q_1 \begin{pmatrix} \Sigma & 0 \end{pmatrix} Q_2$$
with singular values $1 \geq \sigma_1 \geq \sigma_2 \geq ... \geq \sigma_{r_1} \geq 0$. From Proposition 4, $\mathrm{col}(A_1) \cap \mathrm{col}(A_2) = \{0\}$, hence $\sigma_1 < 1$ (Golub and Van Loan, 2012, Theorem 6.4.2). If all $\sigma_i > 0$, then decomposition holds with $N_i = A_i$ and $O_i = \mathbf{0}$. If all $\sigma_i = 0$, then it holds with $O_i = A_i$ and $N_i = \mathbf{0}$. Otherwise, let $k$ denote the minimal $j$ such that $\sigma_j > 0$ so that
$$\sigma_1 \geq \cdots \geq \sigma_k > \sigma_{k+1} = ... = \sigma_{r_1} = 0.$$

Since $Q_1 \in \mathbb{R}^{r_1 \times r_1}$ and $Q_2 \in \mathbb{R}^{r_2 \times r_2}$ are full orthogonal matrices, we can change the basis by taking $B_1 = U_1 Q_1$, $B_2 = U_2 Q_2$ so that $B_1^\top B_2 = (\Sigma\ 0)$. Let $P_B$ denote the projection matrix onto the column space of $B$. Then

$$A_1 = P_{B_1} A_1 = \sum_{i=1}^{k} b_{1i} b_{1i}^\top A_1 + \sum_{i=k+1}^{r_1} b_{1i} b_{1i}^\top A_1 = N_1 + O_1,$$

$$A_2 = P_{B_2} A_2 = \sum_{i=1}^{k} b_{2i} b_{2i}^\top A_2 + \sum_{i=k+1}^{r_2} b_{2i} b_{2i}^\top A_2 = N_2 + O_2.$$

Since $B_1^\top B_2 = \begin{pmatrix} diag(\sigma_1, \ldots, \sigma_k) & 0 \\ 0 & 0 \end{pmatrix}$ with $1 > \sigma_1 \geq \cdots \geq \sigma_k > 0$, it follows that the above decomposition satisfies all the requirements of Proposition.

**Uniqueness.** Assume there exist another $N_i'$, $O_i'$ satisfying the conditions of proposition such that $N_i' \neq N_i$, $O_i' \neq O_i$ from part 1. Let $B_1 \in \mathbb{R}^{n \times r_1}$ and $B_2 \in \mathbb{R}^{n \times r_2}$ be the matrices of orthonormal basis for $A_1$ and $A_2$ from above such that

$$B_1^\top B_2 = \begin{pmatrix} B_{11}^\top \\ B_{12}^\top \end{pmatrix} (B_{21}\ B_{22}) = \begin{pmatrix} diag(\sigma_1, \ldots, \sigma_k) & 0 \\ 0 & 0 \end{pmatrix}, \tag{B.3}$$

where $1 > \sigma_1 \geq ... \geq \sigma_k > 0$. Since $\mathrm{col}(N_i') \perp \mathrm{col}(O_i')$, and $\mathrm{row}(N_i') \cap \mathrm{row}(O_i') = \{0\}$, it follows that $\mathrm{col}(O_i') \subseteq \mathrm{col}(A_i)$, hence
$$O_i' = P_{B_i} O_i' = P_{B_{i1}} O_i' + P_{B_{i2}} O_i'. \tag{B.4}$$
We further show that $P_{B_{11}} O_1' = 0$, the proof for $P_{B_{21}} O_2' = 0$ is analogous.

Conditions 3 and 4 ensure that $\mathrm{col}(O_1') \perp \mathrm{col}(A_2)$, hence
$$0 = P_{B_2} O_1' = P_{B_2} P_{B_{11}} O_1' + P_{B_2} P_{B_{12}} O_1' = B_{21} diag(\sigma_1, \ldots, \sigma_k) B_{11}^\top O_1',$$
where we used (B.3) in the last equality. Since $B_2$ has orthonormal columns, the above equation implies
$$diag(\sigma_1, \ldots, \sigma_k) B_{11}^\top O_1' = 0.$$
Since $\sigma_i > 0$, $i = 1, \ldots, k$, it follows that $B_{11}^\top O_1' = 0$, hence $P_{B_{11}} O_1' = 0$ as claimed. Similarly, one can show $P_{B_{21}} O_2' = 0$. From (B.4), it follows that $\mathrm{col}(O_i') \subseteq \mathrm{col}(B_{i2})$.

Next, we will show that $\text{col}(O_i') = \text{col}(B_{i2})$. Assume this is not true for $i = 1$, that is $\text{col}(O_1') \subset \text{col}(B_{12})$ and $\dim(\text{col}(O_1')) < \dim(\text{col}(B_{12}))$. Since $\text{col}(A_1) = \text{col}(B_{11}) \cup \text{col}(B_{12})$, and $\text{row}(N_1') \cap \text{row}(O_1') = \{0\}$, it follows that $\dim(\text{col}(N_1')) > \dim(\text{col}(B_{11}))$. Since condition 5 implies that $\dim(\text{col}(N_1')) = \dim(\text{col}(N_2'))$, we also must have $\dim(\text{col}(N_2')) > \dim(\text{col}(B_{21}))$, and subsequently $\dim(\text{col}(O_2')) < \dim(\text{col}(B_{22}))$. Let $k = \dim(\text{col}(B_{11})) = \dim(\text{col}(B_{21}))$, then $r = \dim(\text{col}(N_1')) = \dim(\text{col}(N_2')) > k$. Since the maximal principle angle between $\text{col}(N_1')$ and $\text{col}(N_2')$ is strictly smaller than $\pi/2$, we can choose basis $Q_1 \in \mathbb{R}^{n \times r}$ for $N_1$, and $Q_2 \in \mathbb{R}^{n \times r}$ for $N_2$ with $r \geq k + 1$ such that

$$Q_1^\top Q_2 = \Sigma'$$

where $\Sigma'$ is the diagonal matrix with $r \geq k+1$ strictly positive diagonal elements (corresponding to the cosines of principal angles). Using properties 2 and 3, we can chose the basis for $A_1$ as $B_1' = (Q_1, \text{col}(O_1))$ and $B_2' = (Q_2, \text{col}(O_2))$ with the singular value decomposition of $B_1'^\top B_2'$ having $r \geq k+1$ non-zero singular values. However, there exist only $k < r$ non-zero singular values in the SVD of $B_1^\top B_2$. Since the singular values are invariant to the change of basis, we arrived at contradiction, so we must have $r \leq k$, which completes the proof that $\text{col}(O_i') = \text{col}(B_{i2})$ for all $i$. Using property 2 and 3, we conclude that we must also have $\text{col}(N_i') = \text{col}(B_{i1})$. Finally, it follows that $N_i' = N_i$, and $O_i' = O_i$ since

$$N_i' = P_{B_{i1}} A_i = N_i \quad \text{and} \quad O_i' = P_{B_{i2}} A_i = O_i.$$

□

## B.6 Proof of Proposition 6

*Proof.* **Part 1: Existence.** Let $B_1 \in \mathbb{R}^{n \times r}$ and $B_2 \in \mathbb{R}^{n \times r}$ be the matrices of orthonormal basis vectors for $N_1$ and $N_2$ such that $B_1^\top B_2 = \Sigma$, where $\Sigma \in \mathbb{R}^{r \times r}$ is a diagonal matrix of singular values with non-zero elements (note that such matrices $B_1$ and $B_2$ can be constructed following the proof of Proposition 5). Let $P_B$ denote the projection matrix onto the column space of $B$. It follows that

$$N_1 = P_{B_1} N_1, \quad N_2 = P_{B_2} N_2 = P_{B_1} P_{B_2} N_2 + (I - P_{B_1}) P_{B_2} N_2,$$

and matrices $C_1 = P_{B_1} N_1$, $I_1 = \mathbf{0}$, $C_2 = P_{B_1} P_{B_2} N_2$, $I_2 = (I - P_{B_1}) P_{B_2} N_2$ satisfy the requirements of proposition. The decomposition is not unique since exchanging $N_1$ and $N_2$ above leads to $C_1' = P_{B_2} P_{B_1} N_1$, $I_1' = (I - P_{B_2}) P_{B_1} N_1$, $C_2' = P_{B_2}$, $I_2' = \mathbf{0}$, which also satisfy the requirements.

**Part 2: Uniqueness of rank.** Note that $C_1^\top C_2 = N_1^\top N_2$, hence $C_1^\top C_2$ is unique due to uniqueness of $N_1$ and $N_2$. Moreover, property (e) implies

$$\text{rank}([N_1 \ N_2]) = 2\,\text{rank}(N_1) = \text{rank}([C_1 \ C_2]) + \text{rank}(I_1) + \text{rank}(I_2),$$

hence $\dim(\text{col}(I_1)) + \dim(\text{col}(I_2)) = \text{rank}(N_i)$ if $\text{rank}(C_1) = \text{rank}(C_2) = \text{rank}(N_i)$. Therefore, it remains to show $\text{rank}(C_1) = \text{rank}(C_2) = \text{rank}(N_i)$.

If $\text{rank}(C_i) < \text{rank}(N_i)$, then $\text{rank}(C_1^\top C_2) < \text{rank}(N_1^\top N_2)$, which contradicts $N_1^\top N_2 = C_1^\top C_2$. If $\text{rank}(C_i) > \text{rank}(N_i)$, then due to $\text{col}(C_i) = \text{col}(C_1, C_2)$, we must have $\text{rank}(C_1^\top C_2) > \text{rank}(N_i) = \text{rank}(N_1^\top N_2)$, which contradicts $N_1^\top N_2 = C_1^\top C_2$. □

## C Derivation of Algorithm 1

Here we derive the closed-form updates of $V$ and $U$ for solving problem (8).

## C.1 Update of $V$

In this section we derive an algorithm for solving

$$V^{(k)} = \operatorname*{argmin}_{V \in \mathbb{R}^{p \times r}} \left\{ \sum_{i=1}^{d} \left[ \frac{1}{2} \|X_i - U^{(k-1)} V_i^\top\|_F^2 + \lambda \sum_{j=1}^{r} \|V_{ij}\|_2 \right] \right\},$$

where $U^{(k-1)} \in \mathbb{R}^{n \times r}$ has orthonormal columns. For notational simplicity, we drop the subscript and let $U = U^{(k-1)}$. Since the objective function is a sum of functions over $V_i$, $i = 1, \ldots, d$, it follows that optimization with respect to $V_i$ is independent from $V_j$, $j \neq i$. Therefore, we only consider solving

$$V_1^{(k)} = \operatorname*{argmin}_{V_1 \in \mathbb{R}^{p_1 \times r}} \left\{ \frac{1}{2} \|X_1 - U V_1^\top\|_F^2 + \lambda \sum_{j=1}^{r} \|V_{1j}\|_2 \right\}.$$

This is a convex problem, and the solution must satisfy the KKT conditions (Boyd and Vandenberghe, 2004):

$$-X_1^\top U + U^\top U V_1^{(k)} + \lambda \Gamma_1 = 0,$$

where $\Gamma_{1j}$ is the element of subdifferential of $\|V_{1j}^{(k)}\|_2$:

$$\Gamma_{1j} = \begin{cases} V_{1j}^{(k)} / \|V_{1j}^{(k)}\|_2, & V_{1j}^{(k)} \neq 0 \\ \in \{V : \|V\|_2 \leq 1\}, & V_{1j}^{(k)} = 0. \end{cases}$$

Since $U^\top U = I$, it follows that the optimal $V_{1j}^{(k)}$, $j = 1, \ldots, r$, must satisfy

$$V_{1j}^{(k)} = \max\left(0, \left[1 - \frac{\lambda}{\|X_1^\top U_j\|_2}\right]\right) X_1^\top U_j.$$

## C.2 Update of $U$

In this section, we derive the update for

$$U^{(k)} \leftarrow \operatorname*{argmin}_{U : U^\top U = I} \left\{ \frac{1}{2} \sum_{i=1}^{d} \|X_i - U V_i^{(k)\top}\|_F^2 \right\}, \tag{C.5}$$

where $V^{(k)}$ is given. For notational simplicity, we drop the subscript and let $V = V^{(k)}$.

Let $X = [X_1, \ldots, X_d]$ be the concatenated data matrix. Then (C.5) can be rewritten as

$$U^{(k)} \leftarrow \operatorname*{argmin}_{U : U^\top U = I} \left\{ \|X - U V^\top\|_F^2 \right\},$$

which is an orthogonal Procrustes problem (Golub and Van Loan, 2012, Chapter 6.4.1). Therefore, $U^{(k)} = RQ^\top$, where $R$ and $Q$ are from the singular value decomposition of $XV$, $XV = RLQ^\top$.